\documentclass[journal]{IEEEtran}
\usepackage{graphicx} 
\usepackage{amsmath,amssymb,amsbsy,latexsym}
\usepackage[caption=false,font=footnotesize]{subfig}
\usepackage{url}
\usepackage{algorithm}
\usepackage{algpseudocode}

\usepackage{amsmath, amssymb, amsthm}
\usepackage[caption=false,font=footnotesize]{subfig}
\usepackage{array}
\usepackage{makecell}
\usepackage{graphicx}
\usepackage[mathscr]{eucal}
\usepackage{cite}
\usepackage{enumerate}
\usepackage[table]{xcolor}
\usepackage{empheq}  
\usepackage{enumitem}
\setlist{nolistsep}
\IEEEoverridecommandlockouts                              

\newcolumntype{x}[1]{>{\centering\arraybackslash}p{#1}}

\newcommand{\mb}{\mathbb}
\newcommand{\mc}{\mathcal}

\newtheorem{definition}{Definition}

\newif\ifplot
\plottrue

\usepackage{caption}

\usepackage[margin=25pt, labelfont=bf]{caption}
\usepackage{amsmath, amsfonts, pifont, float, color, url}
\usepackage[pdftex, plainpages=false, pdfpagelabels,
bookmarks=true]{hyperref}

\usepackage{aliascnt}
\newaliascnt{eqfloat}{equation}
\newfloat{eqfloat}{h}{eqflts}
\floatname{eqfloat}{Equation}

\newcommand*{\ORGeqfloat}{}
\let\ORGeqfloat\eqfloat
\def\eqfloat{%
  \let\ORIGINALcaption\caption
  \def\caption{%
    \addtocounter{equation}{-1}%
    \ORIGINALcaption
  }%
  \ORGeqfloat
}

\begin{document}

\title{\LARGE \bf
A Deep Learning and Gamification Approach to Energy Conservation at Nanyang Technological University}

\author{\IEEEauthorblockN{Ioannis C.
Konstantakopoulos\thanks{\textsuperscript{*}Corresponding author. email: {\tt ioanniskon@eecs.berkeley.edu}}\textsuperscript{*}, Andrew R. Barkan, Shiying He, Tanya Veeravalli, Huihan Liu, Costas J. Spanos}
\thanks{I. Konstantakopoulos, Shiying He, Tanya Veeravalli, Huihan Liu, Costas J. Spanos are with the
Electrical Engineering and Computer Sciences Department, University of
California, Berkeley, Berkeley, CA 94720.}
\thanks{Andrew R. Barkan is with the Mechanical Engineering Department, University of
California, Berkeley, Berkeley, CA 94720.}
}

\maketitle

\thispagestyle{empty}
\pagestyle{empty}

\begin{abstract}
  The implementation of smart building technology in the form of smart infrastructure applications has great potential to improve sustainability and energy efficiency by leveraging humans-in-the-loop strategy. Adoption of human-centric building services and amenities also leads to improvements in the operational efficiency of cyber-physical systems that are used to control building energy usage. However, human preference in regard to living conditions is usually unknown and heterogeneous in its manifestation as control inputs to a building. Furthermore, the occupants of a building typically lack the independent motivation necessary to contribute to and play a key role in the control of smart building infrastructure. Moreover, true human actions and their integration with sensing/actuation platforms remains unknown to the decision maker tasked with improving operational efficiency. By modeling user interaction as a sequential discrete game between non–cooperative players, we introduce a gamification approach for supporting user engagement and integration in a human-centric cyber-physical system. We propose the design and implementation of a large-scale network game with the goal of improving the energy efficiency of a building through the utilization of cutting-edge Internet of Things (IoT) sensors and cyber-physical systems sensing/actuation platforms. By observing human decision makers and their decision strategies in their operation of building systems, we can apply inverse learning techniques in order to estimate their utility functions. Game theoretic analysis often relies on the assumption that the utility function of each agent is known a priori; however, this assumption usually does not hold in many real-world applications. We propose a benchmark utility learning framework that employs robust estimations for classical discrete choice models provided with high dimensional imbalanced data. To improve forecasting performance, we extend the benchmark utility learning scheme by leveraging Deep Learning end-to-end training with Deep bi-directional Recurrent Neural Networks. We apply the proposed methods to high dimensional data from a social game experiment designed to encourage energy efficient behavior among smart building occupants in Nanyang Technological University (NTU) residential housing. Using occupant-retrieved actions for resources such as lighting and A/C, we simulate the game defined by the estimated utility functions to demonstrate the performance of the proposed methods on ground truth data. For demonstrations of our infrastructure and for downloading de-identified high dimensional data sets, please visit our web site \footnote{\textit{smartNTU} demo web-portal: https://smartntu.eecs.berkeley.edu}.

\end{abstract}


\section{Introduction}
\label{sec:intro}

Nearly half of all energy consumed in the U.S. can be attributed to residential and commercial buildings~\cite{mcquade2009}. In efforts to improve energy efficiency in buildings, researchers and industry leaders have attempted to implement control and automation approaches alongside techniques like incentive design and price adjustment to more effectively regulate energy usage. There has been success in improving building energy efficiency from meter to consumer through the use of model predictive control~\cite{Aswani:2012kx,boman:1998aa,maAnderson2011ACC,lovett:2013aa}).
 
 
Recently, utility companies have invested in demand response programs that can address improper load forecasting while also helping building managers encourage energy efficiency among building occupants~\cite{albadi:2008aa,mathieu:2011aa}. Typically, the implementation of these programs is enacted on a contract basis between utility providers and the consumers under arranged conditions of demand---usage. The building managers will then be bound by contract to operate according to the agreed upon schedule. The setup involves creating a gamification interface to allow building managers to interact with occupants. This interface is designed to support engagement and integration in a human-centric cyber-physical system.

 
Human decision-makers play a critical role in the management and operation of contemporary infrastructure, due in no small part to the advent of Internet of Things and cyber-physical sensing/actuation platforms in industry. The cooperation of human elements with building automation in smart infrastructure helps improve system robustness and sustainability through a combination of control and flexibility. This flexibility makes it possible to accommodate situations like automatic shifting or curtailing demand during peak hours. Put into more broad terms, the goal of many infrastructure systems is to enact system-level efficiency improvements by using a high-level \emph{planner} (e.g. facility manager) to coordinate autonomously acting agents in the system (e.g. selfish human decision-makers). It is this type of functionality that makes smart building technology so essential to the development of an ideal \emph{smart city}.

Our approach to efficient building energy management focuses on residential buildings and utilizes cutting edge Internet of Things (IoT) sensors and cyber-physical systems sensing/actuation platforms. We design a social game aimed at incentivizing occupants to modify their behavior so that the overall energy consumption in their room is reduced. In certain living situations, occupants in residential buildings are not responsible for paying for the energy resources they consume. Hence, there is often an imbalance between the incentives of the facility manager and the occupants. The competitive aspects inherent to the social game framework motivate occupants to address their inefficiencies as well as encourage responsible energy usage on an individual basis.

 
 
At the core of our approach is the decision to model the occupants as non-cooperative agents who play according to a sequential discrete choice game. Discrete choice models have been used extensively to examine modes of transportation~\cite{train1978validation}, choice of airport~\cite{bacsar2004parameterized}, demand for organic foods~\cite{gracia2008demand}, robbery patterns \cite{bernasco2009offenders}, and even school social interactions~\cite{soetevent2007discrete}. Under this assumption of non-cooperation, we were able to use a randomized utility framework and propose novel utility estimation algorithms for the occupants' utility functions. Most importantly, estimating agent utility functions via our method results in predictive models that provide excellent forecasting of occupant actions---usage.

Our framework is centered around learning agent preferences over room resources such as lighting and A/C as well as external parameters like weather conditions, high-level grid control, and provided incentives. Specifically, we model decision making agents as sequential utility maximizers. Agents are strategic entities that make decisions based on their own preferences without consideration of others. The game-theoretic framework both allows for qualitative insights to be made about the outcome of aforementioned selfish behavior---more so than a simple prescriptive model---and, more importantly, can be leveraged in designing mechanisms for incentivizing agents.

 The broader purpose of this paper is to introduce a general framework that leverages game-theoretic concepts to learn models of players' decision making in residential buildings provided with our implementation of a novel energy social game. We present a variety of benchmark utility learning models and a novel pipeline for efficient training. In order to boost predictive power, we propose end-to-end Deep Learning models focused on the utilization of deep recurrent neural networks for capturing gaming data. To make use of sequential information dependencies in our data, we leverage a Deep Learning architecture using Long Short Term Memory cells (LSTM).

Lastly, we provide a demo web portal for demonstrating our infrastructure and for downloading de-identified high dimensional data sets\footnote{\textit{smartNTU} demo web-portal: https://smartntu.eecs.berkeley.edu}. High-dimensional data sets can serve either as a benchmark for discrete choice model learning schemes or as a great source for analyzing occupants' usage in residential buildings. Towards this scope, we use conventional deep variational auto-encoders~\cite{kingma2013auto} or recurrent network based adaptation of variational auto-encoders~\cite{gregor2015draw}  as an approach to create a nonlinear manifold (encoder) that can be used as a generative model. Variational auto-encoders can fit large high dimensional data sets (like our social game application) and train a deep model to generate data like the original data set. In a sense, generative models automate the natural features of a data set and then provide a scalable way to reproduce known data. This capability can be employed either in the utility learning framework for boosting estimation or as a general way to create simulations mimicking occupant behavior---preferences. 

The rest of the paper is organized as follows. Previous works are surveyed in Section 2 with an emphasis on human-centric models. Section 3 describes the social game experimental on the Nanyang Technological University campus and its software architecture setup. In Section 4, we formulate the human decision making model and introduce the core random utility learning framework. Section 5 introduces a novel pipeline for utility estimation along with several proposed machine learning algorithms. Leveraging of deep learning architectures in sequential decision games is proposed in Section 6. Presentation of our results is given in Section 7. We make concluding remarks and discuss future directions along with a brief explanation of our publicly available data set in our demo web portal in Section 8.

\section{Related Work}
\label{sec:rel_work}
Contemporary building energy management techniques employ a variety of algorithms in order to improve performance and sustainability, and many of these approaches leverage ideas from topics such as optimization theory and machine learning. Our goal was to improve building energy efficiency by introducing a gamification system that engages users in the process of energy management and integrates seamlessly through the use of human-centric cyber-physical technology. There exists a considerable amount of previous work on the success of control and automation in the improvement of building energy efficiency~\cite{maAnderson2011ACC,Aswani:2012kx,boman:1998aa}. Some notable techniques with encouraging results implement concepts such as incentive design and adaptive pricing~\cite{Mathieu2012,Dahleh2010smartCom}. Control theory has been a critical source for several approaches that employ ideas like model predictive and distributed control and have demonstrated encouraging results in applications like HVAC. Unfortunately, these control approaches, which are applied to human-centric environments, lack the ability to consider the individual preferences of occupants. This trend is also apparent in machine learning approaches to HVAC system control. While these approaches are capable of generating optimal control designs, they fail to adjust to occupant preferences and the consequences of their presence in the system. The heterogeneity of user preferences in regard to building utilities is considerable in variety and necessitates a system that can adequately account for differences from occupant to occupant.


Clearly, the presence of occupants greatly complicates the determination of an efficient building management system. With this in mind, focus has shifted toward modeling occupant behavior within the system in an effort to incorporate their preferences. To accomplish this task, the building and its occupants are represented as a multi-agent system targeting occupant comfort~\cite{boman:1998aa}. First, occupants and managers are allowed to express their building preferences, and these preferences are used to generate an initial control policy. An iteration on this control policy is created by using a rule engine that attempts to find compromises between preferences. Some drawbacks of this control design are immediately apparent. There should be some form of communication to the manager about occupant preferences. In addition, there exists no incentive for submission of true user preferences, and no system is in place for feedback from occupants. Other related topics in the same vein focus on grid integration~\cite{samad:2016aa} while still others consider approaches for policy recommendations and dynamic pricing systems \cite{Mathieu2012,Dahleh2010smartCom}.


As alluded to previously, the key to our approach is the implementation of a social game among users in a non-cooperative setting. Similar methods that employ \textit{social games} have been applied to transportation systems with the goal of improving flow~\cite{merugu:2009aa,pluntke2013insinc}. Another example application can be found in the medical industry in the context of privacy concerns versus expending calories~\cite{bestick:2013aa}. Entrepreneurial ventures have also sought to implement solutions of their own to the problem of building energy efficiency. \textit{Comfy}\footnote{\tt https://comfyapp.com} and \textit{Cool Choices}\footnote{\tt https://coolchoices.com/how-it-works/improve} are two examples of start-ups that have developed interesting approaches to controlling building utilities. \textit{Comfy}'s product is primarily data-driven, while allowing occupants to have flexibility and independence in their control of HVAC. On the other hand, \textit{Cool Choices} has designed a game theoretic framework centered around improving sustainability. Representing the larger industry agents, companies such as \textit{Oracle - OPower}\footnote{\tt https://opower.com} are attempting to balance energy efficiency and demand response management in an effort to engage customers digitally. Finally, it has been shown that societal network games are useful in a \textit{smart city} context for improving energy efficiency and human awareness~\cite{cowley2011learning,de2014social}.

 
 
 The idea behind the social game context is to create friendly competition between occupants. In turn, this competition will help motivate them to individually consider their own energy usage and, hopefully, seek to improve it. This same technique of gamification has been used as a way to educate the public about energy usage~\cite{knol:2011aa,salvador:2012aa,orland:2014aa}. In previous work, we have explored utility learning and incentive design as a
coupled problem both in theory~\cite{konstantakopoulos2017leveraging, konstantakopoulos2017robust, konstantakopoulos:2016ab, konstantakopoulos:2016aa, jin2015rest, jia2018poisoning} and in
practice~\cite{konstantakopoulos2014social, ratliff2014social, ratliff:2014ac, jin:2015aa} under a Nash equilibrium approach. Gamification approaches are presented in platform-based design flow for smart buildings~\cite{jia2018design}. It has also been cleverly implemented in a system that presents feedback about overall energy consumption to occupants~\cite{simon:2012aa}. One case of a gamification methodology was used to engage individuals in demand response (DR) schemes~\cite{li:2014aa}. Each of the users is represented as a utility maximizer within the model of a Nash equilibrium where occupants gain incentive for reduction in consumption during DR events. In contrast to approaches that target user devices with known usage patterns~\cite{li:2014aa}, our approach focuses on personal room utilities such as lighting without initial usage information, simulating scenarios of complete ignorance to occupant behaviors. For our method, we leverage past user observations to learn the utility functions of individual occupants by the way of several novel algorithms. Through this approach, we can generate excellent prediction of expected occupant actions. Our unique social game methodology simultaneously learns occupant preferences while also opening avenues for feedback. This feedback is translated through individual surveys that provide opportunities to influence occupant behavior with adaptive incentive. With this technique, we are capable of accommodating occupant behavior in the automation of building energy usage by learning occupant preferences and applying a variety of novel algorithms. Furthermore, the learned preferences can be adjusted through incentive mechanisms to enact improved energy usage.

A series of experimental trials were conducted to generate real-world data, which was then used as the main source of data for our approach. This differentiates our work from a large portion of other works in the same field that use simulations in lieu of experimental methods. In many cases, participants exhibit a tendency to revert to previously inefficient behavior after the conclusion of a program. Our approach combats this effect by implementing incentive design that can adapt to the behavior and preferences of occupants progressively, which ensures that participants are continuously engaged. From a managerial perspective, the goal is to minimize energy consumption while maximizing occupant comfort. With this social game framework, the manager is capable of considering the individual preferences of the occupants within the scope of the building's energy consumption. The advent of this social game system could potentially offer an unprecedented amount of control for managers without sacrificing occupant comfort and independence.


\section{Smart Building Social Game}
\label{sec:games}

\begin{figure}[!ht]
    \subfloat[Graphical user interface (GUI) for energy-based social game: Display of consumption for resources, real-time feedback for device status (on/off), daily counter, and random coin]{%
    \includegraphics[width=0.45\textwidth]{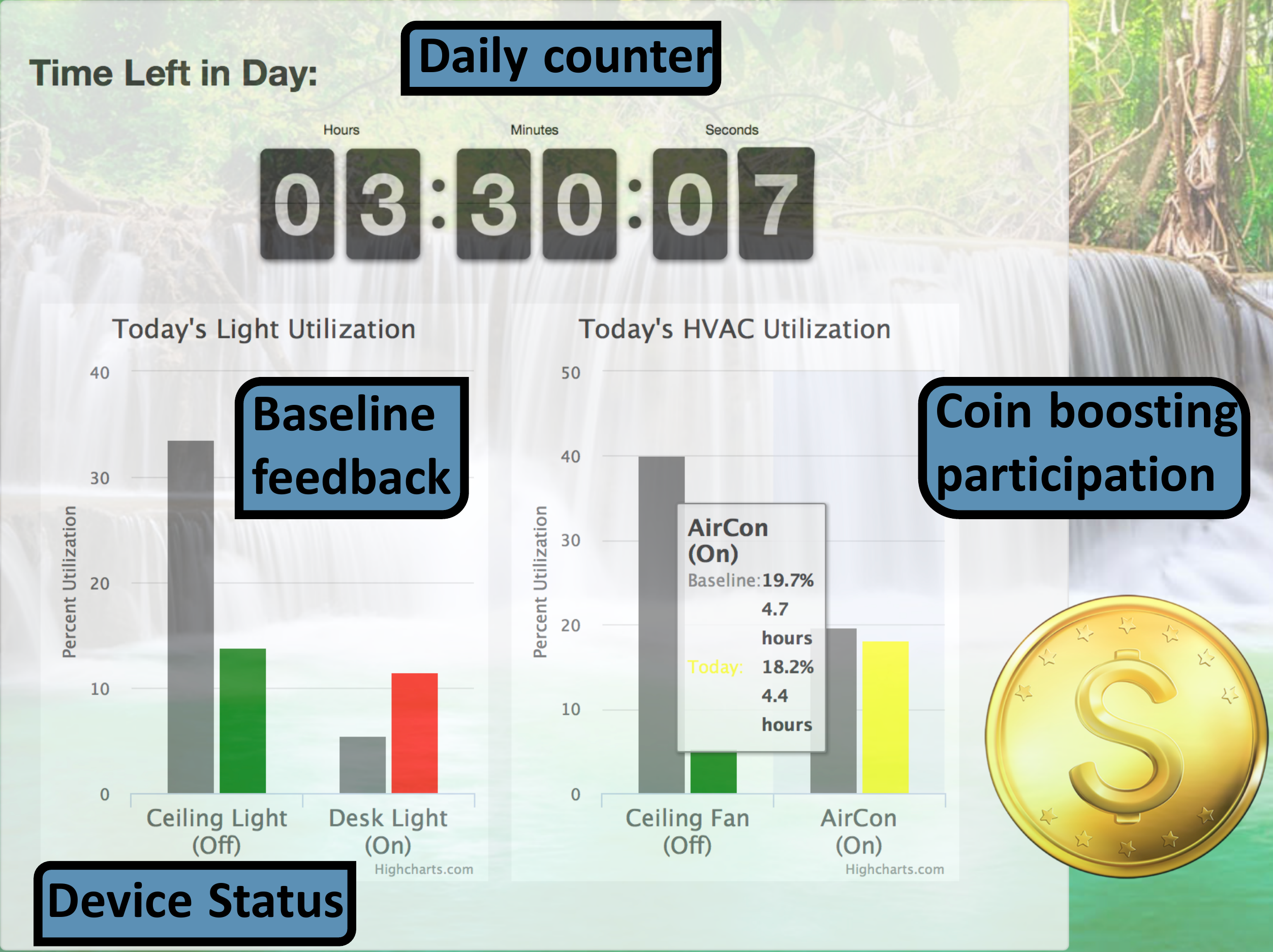}
    }
    \hfill
    \subfloat[Social game dataflow architecture design]{%
    \includegraphics[width=0.46\textwidth]{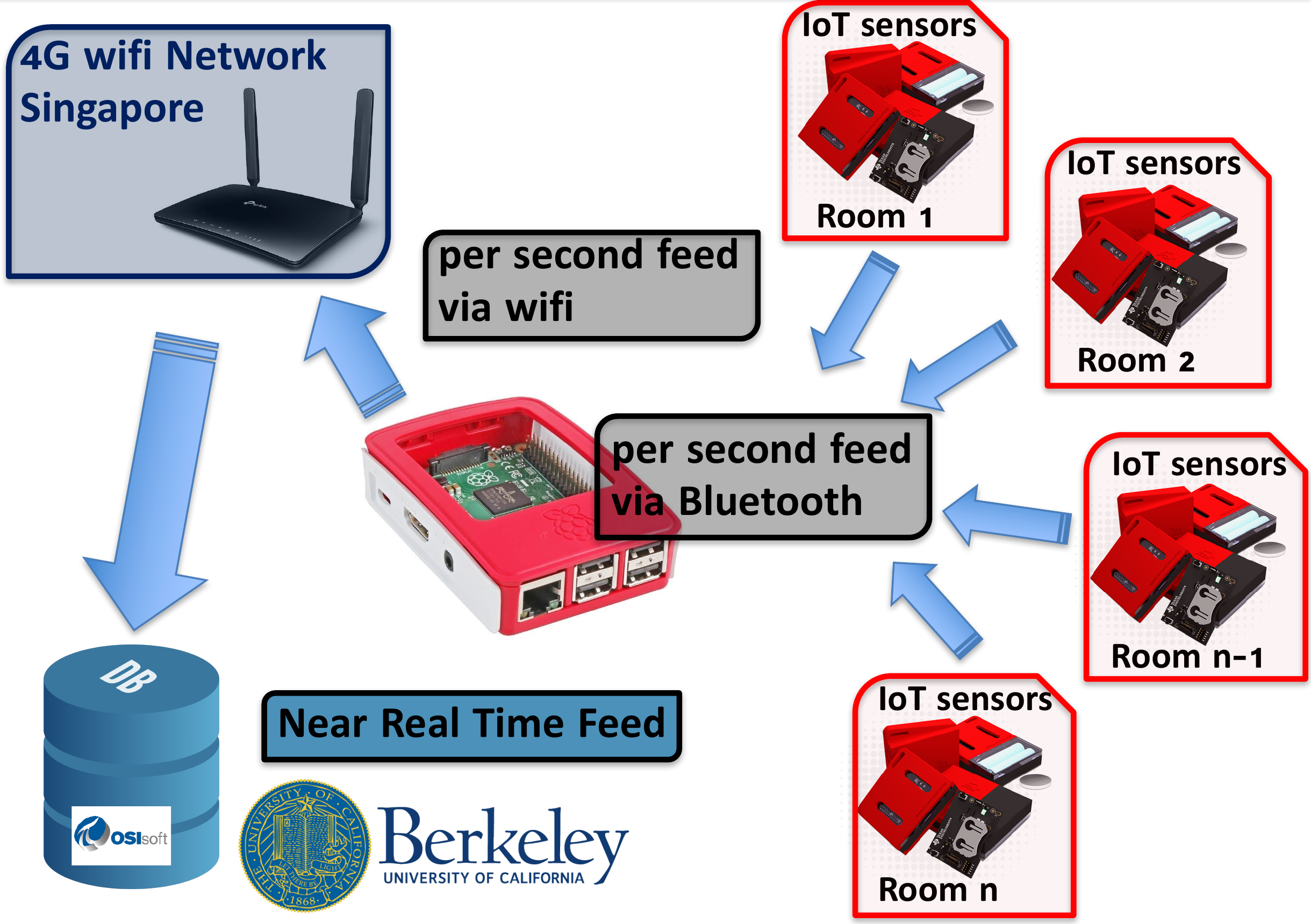}
    }
\caption{Graphical user interface (GUI) and dataflow design for energy-based social game}
\label{fig:game_utilization_design}
\end{figure} 

In this section, we introduce the concept of our social game implemented at Nanyang Technological University (NTU) residential housing apartments along with the software architecture design for the deployed Internet of Things (IoT) sensors.

\subsection{Description of the Social Game Experiment}

Our experimental environment is comprised of residential housing single room apartments on the Nanyang Technological University campus. We designed a social game such that all single room dorm occupants could freely view their daily room's resource usage with a convenient interface. In each dorm room we installed two Internet of Things (IoT) sensors \footnote{\textit{IoT sensor Tag}: http://www.ti.com}---one close to the desk light and another near the ceiling fan. With the deployment of IoT sensors, dorm occupants can monitor in real-time their room's lighting system (desk and ceiling light usage) and HVAC (ceiling fan and A/C usage) with a refresh interval of up to one second. 

Dorm occupants are rewarded with points based on how energy efficient their daily usage is in comparison to their past usage before the social game was deployed. The past usage data that serves as our baseline is gathered by monitoring occupant energy usage for approximately one month before the introduction of the game for each semester. Using this prior data, we calculated a weekday and weekend baseline for each of an occupant's resources. We accumulate data separately for weekdays and weekends so as to maintain fairness for occupants who have alternative schedules of occupancy (e.g. those who tend to stay at their dorm room over the weekends versus weekdays). We employ a lottery mechanism consisting of several gift cards awarded on a bi-weekly basis to incentivize occupants; occupants with more points are more likely to win the lottery. Earned points for each resource is given by

\begin{equation}
  \hat{p}^{d}_{i}(b_i, u_{i})= s_{i} \frac{b_i - u^{d}_{i}}{b_i}
  \label{eq:points_earned}
\end{equation}

where $\hat{p}^{d}_{i}$ is the points earned at day $d$ for room's resource $i$ which corresponds to ceiling light, desk light, ceiling fan, and A/C. Also, $b_i$ is the baseline calculated for each resource $i$, $u^{d}_{i}$ is the usage of the resource at day $d$, and $s_{i}$ is a points booster for inflating the points as a process of framing \cite{tversky1981framing}. This process of framing can greatly impact a user's participation, and it is routinely used in rewards programs for credit cards among many other point-based programs used in industry applications. In addition, we rewarded dorm occupants for the percentage of savings \eqref{eq:points_earned} because we felt it was important to motivate all of the participants to optimize their usage independent of the total amount of energy consumed in their normal schedule. However, over-consumption resulted in negative points.


In Figure~\ref{fig:game_utilization_design}, we present how our graphical user interface was capable of reporting to occupants the real-time status (on/off) of their devices, their accumulated daily usage, time left for achieving daily baseline, and the percentage of allowed baseline being used by hovering above their utilization bars. In order to boost participation, we introduced a randomly appearing coin close to the utilization bars with the purpose of incentivizing occupants to log in to web-portal and view their usage. The coin was designed to motivate occupants towards viewing their resource usage and understanding their impact to energy consumption by getting exact usage feedback in real-time. Based on this game principle, we gave occupants points when they clicked on the coin, which could increase both their perceived and actual chances of winning the rewards. 

The residential housing single room apartments on the Nanyang Technological University campus are divided into four blocks, each of which having two levels. In this space, there is a total of seventy-two occupants who are eligible to participate in the social game. Participation in our social game platform was voluntary. We ran the experiment in both the Fall 2017 (September 12th - December 3rd) and Spring 2018 (February 19th - May 6th) semesters. In the Fall 2017 version, we included ceiling light, desk light, and ceiling fan resources in the graphical user interface for the social game while in the Spring 2018 version we included all of the potential resources that were available.

\subsection{Internet of Things (IoT) System Architecture}

We enabled the design and implementation of a large-scale networked social game through the utilization of cutting-edge Internet of Things (IoT) sensors. In total, we have deployed one hundred and forty-four sensors in dorm single rooms. These are part of a hardware and software infrastructure that achieves a near real-time monitoring of several metrics of resource usage in each room like lighting and A/C, as well as recording detailed indoor conditions. Moreover, our system is capable of saving occupant actions in the web-portal as well as gathering weather data from our installed local weather monitoring station, which acts as an external parameter for our model. Weather data is gathered from an externally-installed local weather monitoring station at per-second resolution.

The actual design and dataflow is depicted in Figure~\ref{fig:game_utilization_design}. IoT sensors' software (firmware) has been updated for achieving highest reliability, continuous posting of indoor conditions in dorm rooms, and optimization of battery usage. In addition, we have designed and printed our own 3D plastic case for securing IoT sensors and for adding larger capacity battery (3V) for longer lasting capabilities. Our deployed IoT sensors are capable of continuously working for six months with a newly installed CR123 lithium battery. The IoT sensors continuously feed data every ten seconds via Bluetooth connection to a nearby deployed Raspberry Pi. To increase the system's reliability, we installed one Raspberry Pi in every other single dorm room. Next, each Raspberry Pi feeds the collected data via Wi-Fi to our deployed Wi-Fi routers connected to the Internet via 4G cellular modems. We have deployed our own 4G Wi-Fi Network since Nanyang Technological University residential housing doesn't have stable campus-wide Wi-Fi. In this way, our system can be deployed and operate independent of the condition of on-site utilities such as Wi-Fi. After receiving the Bluetooth feed, the data is posted to our OSIsoft PI database \footnote{\textit{OSIsoft PI database}: https://www.osisoft.com} located at the University of California, Berkeley campus. 

\begin{figure}[!ht]
    \subfloat[Ceiling fan device status pattern]{%
    \includegraphics[width=0.45\textwidth]{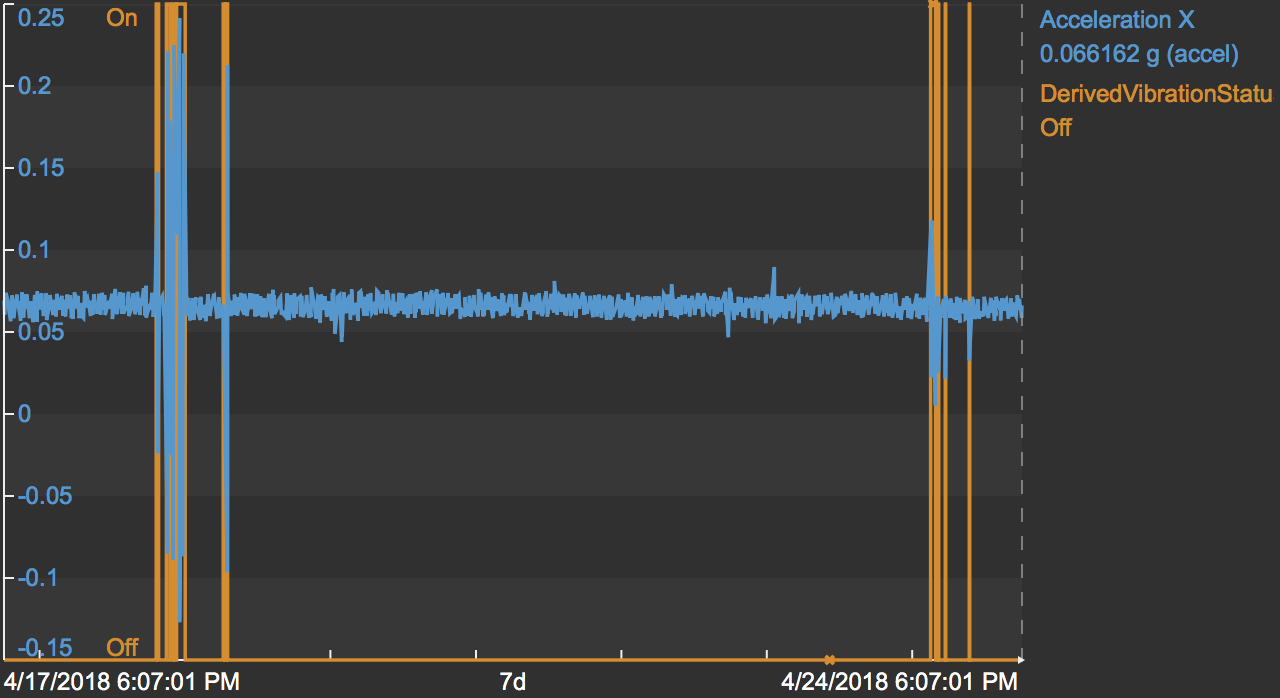}
    }
    \hfill
    \subfloat[Ceiling light device status pattern]{%
    \includegraphics[width=0.446\textwidth]{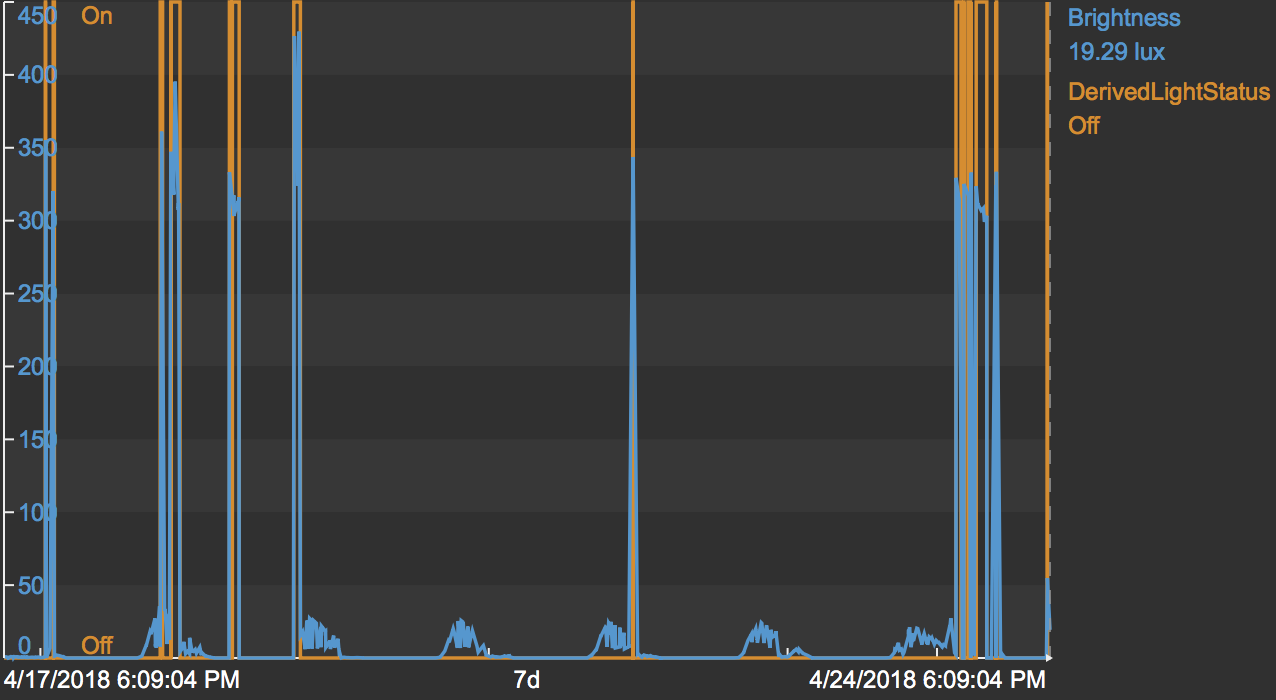}
    }
    \hfill
    \subfloat[A/C device status pattern]{%
    \includegraphics[width=0.45\textwidth]{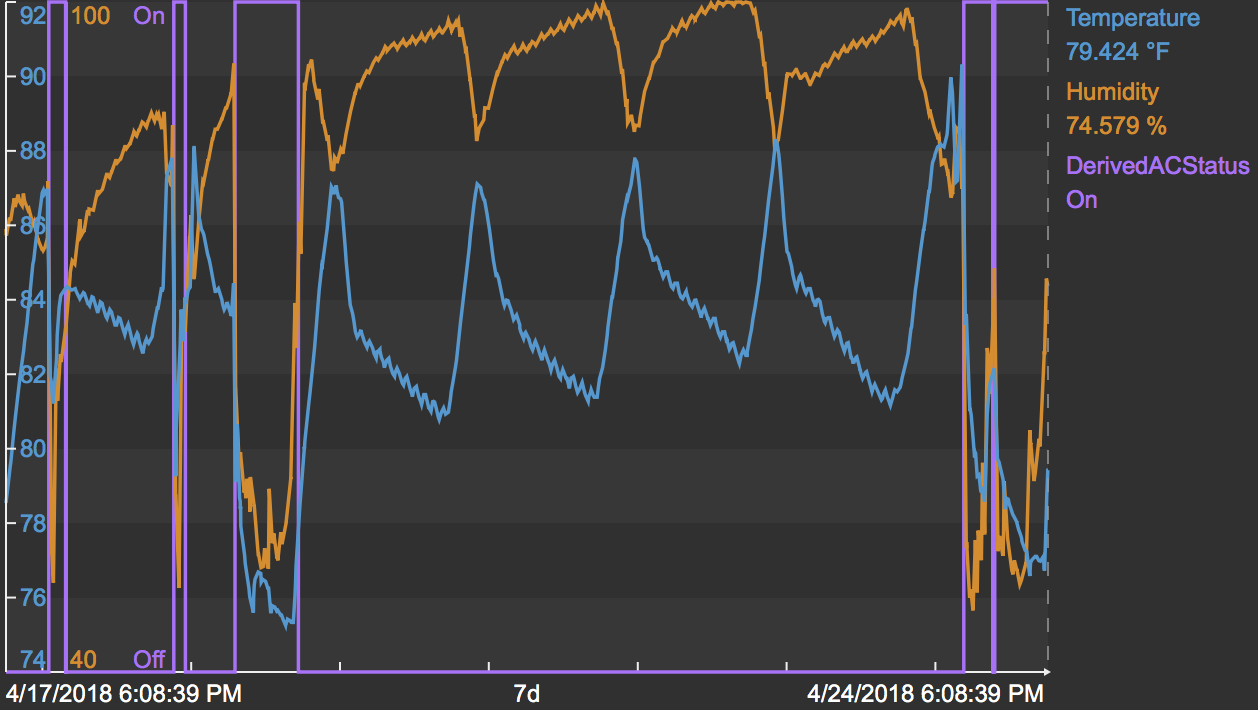}
    }
    \hfill
    \subfloat[Desk light device status pattern]{%
    \includegraphics[width=0.45\textwidth]{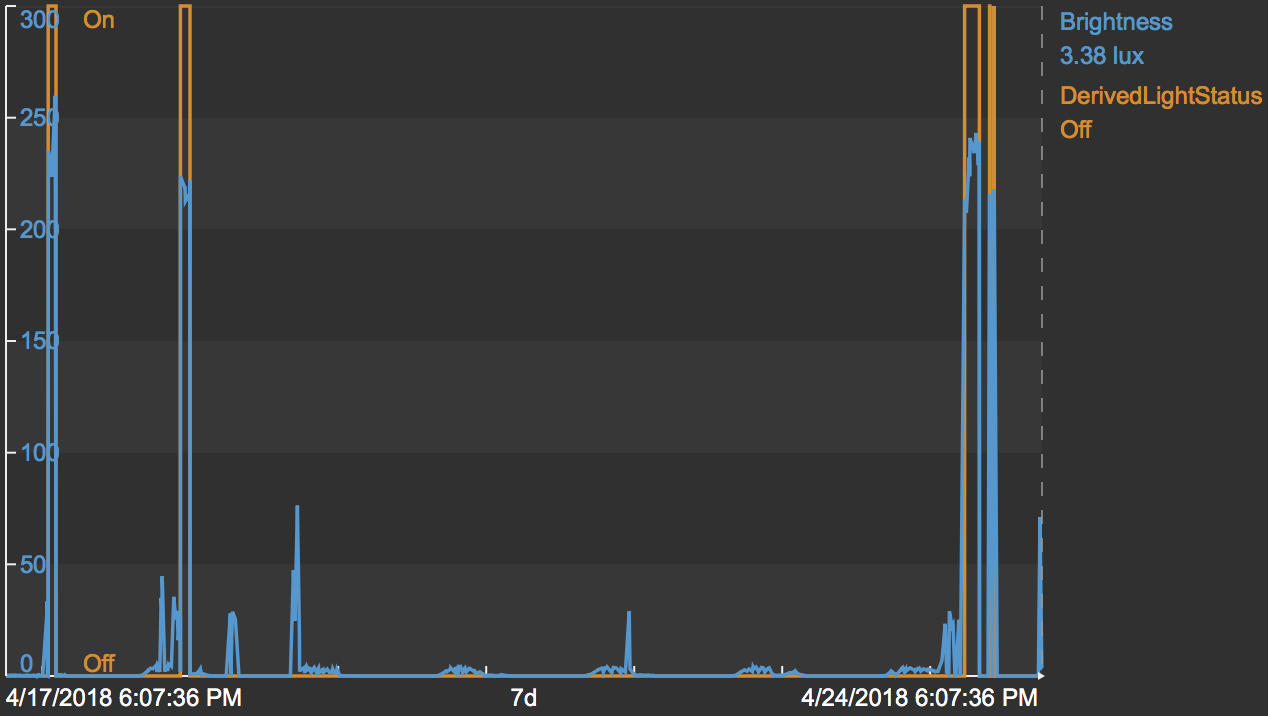}
    }
\caption{Patterns among all targeted resources. The on pulses point to instances that the thresholding system indicates activity in the device.}
\label{fig:device_status}
\end{figure}

Utilizing the data gathered from each dorm room, we leveraged several indoor metrics like indoor illuminance, humidity, temperature, and vibrations for the ceiling fan sensor. Having performed various tests during Summer 2017 within the actual unoccupied dorm rooms, we have derived simple thresholds indicating if a resource is in use or not. For instance, the standard deviation of acceleration derived from the ceiling fan mounted sensor is an easy way to determine whether ceiling fan is in the on state. Additionally, by combining humidity and temperature values, we were able to reliably identify whether A/C is in use with limited false positives. In Figure~\ref{fig:device_status}, we have included row (per second) data and their trends when indoor resources were on/off. Our calibrated detection thresholds are robust over daylight patterns, external humidity/temperature patterns, and noisy measurements naturally acquired from IoT sensors.


While we were getting streaming data from various sensors in all dorm rooms, our back-end processes updated the status of the devices in near real-time in each occupant's account and updated points based on their usage and point formula \eqref{eq:points_earned}. This functionality allows occupants to receive feedback for their actions and view their points balance and their ranking among other capabilities. In order to allow participants to assess and visualize their energy efficient behavior, each user's background in the web-portal changes based on their ranking and energy efficiency. We used background pictures of rain forest settings for encouraging the more energy efficient occupants and images of desert scenes to motivate those with limited energy savings. For a live view of our web portal \footnote{\textit{smartNTU} demo web-portal: https://smartntu.eecs.berkeley.edu}, you can visit our demo web-site, which serves as a demonstration of the game and as a hub for downloading de-identified per-minute aggregated data. The entire web portal and background processes were developed using Django, a python based web development framework. MySQL was used as a database back-end for game related data not stored in the OSIsoft time series Pi database.


\subsection{Social Game Data Det}\label{social_game_data_set}

As a final step, we aggregate occupants' data in per-minute resolution. We have several per-minute features like time-stamp, each resource's status, accumulated resource usage (in minutes) per day, resource baseline, gathered points (both from game and surveys), occupant rank in the game over time, and number of occupant's visits to the web portal. In addition to these features, we add several external weather metrics like humidity, temperature, and solar radiation among others.

After gathering a high dimensional data set with all of the available features, we propose a \textbf{pooling \& picking} scheme to enlarge the feature space and then apply a \textbf{Minimum Redundancy and Maximum Relevance (mRMR)}~\cite{peng2005feature} feature selection procedure to identify useful features for our predictive algorithms. We pool more features utilizing a subset of the already derived features by leveraging domain knowledge and external information. Specifically, we consider: 
  
\begin{enumerate}
    \item \textbf{College schedule dummy feature indicators:} including dummy variables for dates related to breaks, holidays, midterm period, and the final exam schedule at Nanyang Technological University. These features can capture occupants' variability of normal usage of their dorm room resources.
    \item \textbf{Seasonal dummy feature indicators:} we compute several seasonal dummy variables utilizing our time-stamps. Examples of seasonal variables are time of day (morning vs. evening) and time of week (weekday vs. weekend) indicators. The intuition behind such features is their ability to capture occupants' seasonal patterns in resource usage. 
    \item \textbf{Resources status continuous features:} we incorporate stream resource status data for defining pooled features that accurately model occupants variability in resources usage across each day. Examples of such pooled features are: frequency of resource daily switches and percentage of resource usage across the day. 
\end{enumerate}

For more details related to our data set and feature space, visit our web site \footnote{\textit{smartNTU} demo web-portal: https://smartntu.eecs.berkeley.edu}, which includes detailed descriptions of our features.

\section{Human Decision Making: Game Theoretic Framework}
\label{sec:decision_model}
In this section, we abstract the agents’ decision making processes in a game-theoretic framework and introduce discrete choice theory. Discrete choice theory is greatly celebrated in the literature as a means of data-driven analysis of human decision making. Under a discrete choice model, the possible outcome of an agent can be predicted from a given choice set using a variety of available features describing either external parameters or characteristics of the agent. We use a discrete choice model as a core abstraction for describing occupant actions related to their dorm room resources.  

\subsection{Agent Decision Making Model}

Consider an agent $i$ and the decision-making choice set which is mutually exclusive and exhaustive. The decision-making choice set is indexed by the set $\mc{I}=\{\mc{J}^1,\ldots, \mc{J}^S \}$. Decision maker $i$ chooses between $S$ alternative choices and would earn a \textbf{representative utility} $f_i$ for $i \in \mc{I}$. Each decision among decision-making choice set leads agents to get the highest possible utility, $f_i > f_j$ for all $i,j \in \mc{I}$. In our setting, an agent has a utility which depends on a number of features $x_z$ for $z=1, \ldots, N$. However, there are several unobserved components --- features of the representative utility which should be treated as random variables. Hence, we define a \textbf{random utility} decision-making model for each agent given by 

\begin{equation}
  \hat{f}_i(x)=g_i(\beta_{i},x) + \epsilon_{i}
  \label{eq:discrete_utility}
\end{equation}

where $\epsilon_{i}$ is the unobserved random component of the agent's utility, $g_i(\beta_{i},x)$ is a nonlinear generalization of agent $i$'s utility function,
and where

\begin{equation}
x=(x_1, \ldots, x_{i-1}, x_{i+1}, \ldots, x_N)\in \mb{R}^{N} 
\label{eq:feature}
\end{equation} 

is the
collective $n$ features explaining an agent's decision process. The choice of nonlinear mapping $g_i$ and $x$ abstracts the agent's decision; it could represent, e.g.,
how much of a particular resource they choose to use and when an agent optimizes its usage over a specific resource. In general, each agent is modeled as a \emph{utility maximizer} that seeks to select $i\in
\mc{I}$ by optimizing \eqref{eq:discrete_utility}. 

Discrete choice models in their classical representation~\cite{train2009discrete} are given by a linear mapping $g_i(\beta_{i},x) = \beta_{i}^{T}x$ in which $\epsilon_{i}$ is an independently and identically distributed random value modeled using a Gumbel distribution. According to \cite[Chapter~3]{train2009discrete}, the probability that agent $i$ chooses choice $j \in \mc{J}$ is given by

\begin{equation}
P_{i}^{j} = P(\beta_{j}^{T}x + \epsilon_{j} > \beta_{z}^{T}x + \epsilon_{z}, \forall j \neq z\Longrightarrow 
P_{i}^{j} = \frac{\exp{\beta_{j}^{T}x}}{\sum^{\mc{J}}_{s=1} \exp{\beta_{z}^{T}x}}
\label{eq:Logit}
\end{equation}

According to~\eqref{eq:Logit}, each agent's probability of a specific choice is given by a Logit model from the linearity assumption for the feature representative utility and the Gumbel distribution modeling the unknown random variable. Other distributions could be used (e.g. Gaussian for Probit model), and this is a design flexibility of discrete choice models. 

\subsection{Game Formulation}

To model the outcome of the strategic
interactions of agents in the deployed social game, we use a \emph{sequential non-cooperative discrete game} concept. Introducing a generalized decision-making model for each agent~\eqref{eq:discrete_utility}, in which random utility can be modeled either with linear or nonlinear mapping, a sequential non-cooperative discrete game is given by

\begin{definition}

Each agent $i$ has a set $\mc{F}_i = {f_{i}^1,\ldots, f_{i}^{N}}$ of $N$ \textbf{random utilities}. Each random utility $j$ has a convex decision-making choice set $\mc{I}_{j}=\{\mc{J}^{1}_{j},\ldots, \mc{J}^{S}_{j} \}$. Given a collective of $n$ features~\eqref{eq:feature} comprising the decision process, agent $i$ faces the following optimization problem for their \textbf{aggregated random utilities}:

\begin{equation}
  \max\{\sum_{i=1}^{N}f_i(x), \ \text{for} \ f_i \in \mc{F}_i\}.
  \label{eq:opt-seq}
\end{equation}
\end{definition}

In the sequential equilibrium concept, we simulate the game defined by the estimated random utility functions per resource to demonstrate the actual decision-making process of each individual dorm occupant. Agents in the game independently co-optimize their aggregated random utilities~\eqref{eq:opt-seq} given a collective of $n$ features~\eqref{eq:feature} at each time instance. A general incentive design mechanism~\eqref{eq:points_earned} motivates their potential actions across various given decision-making choice sets. 

The above definition extends the definition of a discrete choice model~\cite{train2009discrete} to sequential games in which agents concurrently co-optimize several discrete (usually mutually exclusive) choices. Using this definition, we can apply the proposed game theoretic model by allowing several machine learning algorithms to be directly applied. Machine learning algorithms can potentially be used for modeling the choice of nonlinear mapping~\eqref{eq:discrete_utility}. In particular, Deep Learning models can perform an end-to-end training for higher predictive accuracy using several mini-batched collectives of features~\eqref{eq:feature}.

\section{Benchmark Learning Framework}
\label{sec:benchmark}
In the previous section, we introduced an extension to discrete choice models for sequential decision making over a set of different (usually mutually exclusive) choices. With appropriate modeling of unobserved random components and a linearity assumption for the features mapping in an agents utility function, we have a Logit model~\eqref{eq:Logit}. However, unobserved random components and the overall structure of random utility~\eqref{eq:discrete_utility} can be modeled by a variety of classification machine learning models---either discriminative types like the Logit model or even with generative models like Bayesian Networks. 

In the current section, we examine the utility learning problem using a novel pipeline including a variety of proposed statistical learning methods and models that serve to improve estimation and prediction accuracy for our proposed sequential discrete choice model. The proposed benchmark learning framework scheme can be folded into an overall incentive design framework by either a building manager or utility companies. This goal motivates why we are interested in learning more than a simple predictive model for agents, but rather an exceptional utility-based forecasting framework that accounts for occupants' preferences. Furthermore, well-trained classification models serve as an excellent benchmark for our proposed Deep Learning models or other more advanced and complex sequential learning techniques like Hidden Markov Models or Conditional Random Fields. 

\subsection{Random Utility Estimation Pipeline}

We start by describing the basic components of our proposed random utility estimation pipeline using observed pooled features and data derived from the game played between the agents. The utility learning framework we propose is quite broad in that it encompasses a wide class of discrete choice games, as our proposed game~\eqref{eq:opt-seq} is a super-set containing classical discrete choice models. Let us introduce the pipeline formulation as it serves as the basis for the random utility estimation method.



After gathering streaming data in our MySQL data-base (as described in Section~\ref{social_game}), we pool several candidate features and expand our feature space. Next, a large set of proposed high dimensional candidate features is constructed. Using this feature set, we adopt a greedy feature selection algorithm called Minimum Redundancy Maximum Relevance (mRMR)~\cite{peng2005feature}. The mRMR greedy heuristic algorithm utilizes mutual information as the metric of goodness for a candidate feature set. Given the large number of pooled candidate features, mRMR feature selection is a useful method of finding a subset of features that are relevant for the prediction of occupants' resource usage. The algorithm resolves the trade-off between relevancy and redundancy of the derived feature set by simultaneously reducing redundancy in the features and selecting those most relevant to occupants' actual actions over time. Our target is to avoid adding redundant features that do not boost prediction (e.g. classification) accuracy, but instead cause extra learning noise and increase computation time.

The mRMR feature selection algorithm is applied to batched gathered data from the first game period either in the Fall or Spring version of the Social Game. From the total number of available feature candidates, we decided to keep nearly half of them. Running the mRMR feature selection algorithm using the pooled features for the Fall semester --- with the dorm room ceiling fan being the target resource --- yields results that make logical sense based on the context of the features being examined. Unsurprisingly, ceiling fan percentage of usage is the most relevant and least redundant feature while external humidity seems to be the second most important feature influencing an occupant's ceiling fan usage. This leads us to the conclusion that mRMR is a suitable solution for obtaining the most relevant features in the proposed novel pipeline for utility learning.


After getting a number of top performing features as a result of the mRMR greedy algorithm, we apply a simple data pre-processing step with mean subtraction --- subtracting the mean across each individual feature. Mean subtraction centers the cloud of data around the origin along every dimension. On top of mean subtraction, we normalize the data dimensions by dividing each dimension by its standard deviation in order to achieve nearly identical scale in the data dimensions. However, the training phase of the random utility estimation pipeline has one potentially significant challenge, which is the fact that data in almost every resource data set is heavily imbalanced (e.g. the number of resources with 'off' samples is on the order of 10-20 times more than those with 'on' samples). This is normal considering occupants' daily patterns of resource usage in buildings, but it poses a risk of having potentially poorly trained random utility estimation models.

For optimizing around highly imbalanced data sets, we adapt the Synthetic Minority Over-Sampling (SMOTE)
~\cite{chawla2002smote} technique for providing balanced data sets for each resource and for boosting prediction (e.g. classification) accuracy. SMOTE over-samples a data set used in a classification problem by considering k nearest neighbors (in feature space) of the minority class given one current data point of this class. The SMOTE technique creates several artificial data points by randomly considering a vector along the resulting k neighbors with respect to a current data point. Then, this randomly chosen vector is multiplied by a random number $s \in [0,1]$ and added to the current data point. The outcome vector of this procedure results in newly synthesized data points. The SMOTE algorithm can be initialized by leveraging a pre-processing phase with Support Vector Machines as a grouping step.

After the SMOTE step, we train several classifiers to model~\eqref{eq:discrete_utility} as a final step for the random utility estimation pipeline. These proposed machine learning algorithms do not require strong assumptions about the data process. Moreover, we propose a base model of logistic regression~\eqref{eq:Logit}. In an effort to improve the base discrete choice model, we include penalized logistic regression (regLR) trying $l1$ norm protocol (Lasso) for the model training optimization procedure among other classical classification machine learning algorithms. We perform a randomized grid search for optimizing classifiers using the Area Under the Curve (AUC) metric~\cite{fawcett2006introduction} aiming to co-optimize TPR (sensitivity) and FPR (1-specificity).


We use the Area Under the Receiver Operating Characteristic (ROC) Curve as our performance metric. ROC curves describe the predictive behavior of a binary classifier by concurrently plotting the probability of true positive rate (i.e. correct classification of samples as positive) over false positive rate (i.e. the probability of falsely classifying samples as positive). Using the AUC metric from the ROC curve, we can quantify performance using a single metric to estimate the predictive accuracy of our proposed machine learning classification models as random utility estimators. For training the proposed machine learning algorithms, we used k-fold cross validation combined with the AUC metric to randomly split the data into training and validation sets in order to quantify the performance of each proposed machine learning model in the training phase. We applied 10-fold cross validation, and for each machine learning model we tuned the output model based on those parameters from a random grid search that achieved the best predictive performance on the cross validation set.

Each machine learning algorithm used in our benchmark pipeline and their respective hyper-parameters are described below

\begin{enumerate}
    \item \textbf{Logistic Regression (LR):} Logistic regression is the base model for discrete choice models~\eqref{eq:Logit}. Using a simple sigmoid function, it combines a set of linear features for achieving a posterior classification distribution. 
    \item \textbf{Penalized Logistic Regression (penLR):} Penalized logistic regression combines the cost function of classical Logistic regression with either $l2$ norm (Ridge) or $l1$ norm (Lasso) as a penalty term in the optimization procedure. Ridge and Lasso shrink or control the resulting weights in the obtained model. Lasso tends to result in more sparse models in which several weights can be zero or very close to zero~\cite{Hastie09}. Both Ridge and Lasso penalized logistic regression models are controlled by the $\lambda$ hyper-parameter, which adjusts the penalty term in the optimization scheme.
    \item \textbf{Bagged Logistic Regression (bagLR):} Bagging~\cite{Hastie09} is a powerful ensemble method for combining several weak classifiers and for building a prediction model in which majority vote scheme is applied. Bagging is a technique for reducing the overall variance in the resulting machine learning model. It works based on a bootstrapping technique for re-sampling with replacement for $N$ replicates of the original training data. Then we train $N$ different logistic regression models and combine the resulting bootstrapped estimators by majority vote scheme. There is no immediate hyper-parameter for bagging other than selecting the number of models in which the bagged model will be constituted. 
    \item \textbf{Linear Discriminant Analysis (LDA):} The linear discriminant analysis classification algorithm can be considered as an alternative version of the base model for discrete choice models~\eqref{eq:Logit}. The main difference lies in the fact that the LDA classifier infers joint probability distribution of the decision making features and target (mutual exclusive) values. It defines a prior over the frequency of target values. It also fits a multivariable Gaussian distribution for the modeling of a conditional distribution given an individual target's values. Usually, it gives similar results to Logistic Regression (LR) despite its stronger modeling assumptions. 
    \item \textbf{k-Nearest Neighbors (kNN):} k-Nearest Neighbors classification algorithm is a non-parametric method which finds the k closest resulting training data points sampled across the whole of a given feature space~\cite{Hastie09}. The number of neighbors $k$ is a hyper-parameter which largely controls this classifier's performance. For the k-Nearest Neighbors classifier case, the output given a testing sample is a class member obtained by a majority vote scheme against its neighbors. The k-Nearest Neighbors classification model is easy to build with a fast training phase. However, it has the requirement of caching training data points. Furthermore, it has a computationally expensive testing phase, which makes it impractical in real application settings. 
    \item \textbf{Support Vector Machine (SVM):} The support vector machine model obtains optimal hyperplane(s), which achieves a maximum separation margin between data classes~\cite{Hastie09}. The support vector machine model is controlled by the $C$ hyper-parameter, which adjusts the separation of hyperplane(s) using maximum margin. Moreover, the support vector machine model could combine either $l2$ norm (Ridge) or $l1$ norm (Lasso) as an additional penalty term. The $\lambda$ hyper-parameter again adjusts the penalty term in the optimization scheme. 
    \item \textbf{Random Forest:} Random Forest~\cite{Hastie09} is an efficient ensemble learning method used for classification among other machine learning tasks. It is a variation of classical bagging over decision trees. The Random Forest classifier constructs a large set of decision trees using classical CART greedy algorithm on a randomized subset of given input features. The randomization of input features allows significant reduction of variance in the generalization error. The output of the Random Forest classifier is given by the majority vote of the set of decision trees. One of the advantages of a Random Forest classifier is that testing a new data point takes $O(N*log(h))$ computation time, where $N$ is the number of decision trees and $h$ the height of each decision tree, which happens to be highly balanced. Lastly, Random Forest has several hyper-parameters such as number of decision trees in the model, subset of features to include in the training phase, and depth of each binary trees among others. One significant note is the fact that a potentially large number of used decision trees somehow seem not to lead to over-fitting mainly due to randomization of feature set~\cite{Hastie09}. 
\end{enumerate}

\section{Leveraging Deep Learning for Sequential Decision Making}
\label{sec:deaplearning}
Let us now formulate a novel Deep Learning framework for random utility estimation that allows us to drastically reduce our forecasting error by increasing model capacity and by structuring intelligent deep sequential classifiers. The architecture for deep networks is adaptive to proposed sequential non-cooperative discrete game models and achieves a tremendous increase to forecasting accuracy. Hence, deep networks achieve an end-to-end training for modeling agents' random utility~\eqref{eq:discrete_utility} with extraordinary accuracy.  

The primal version of Artificial Neural Networks (ANNs) can be dated back to 1943~\cite{mcculloch1943logical}. After the emergence of Deep Learning, several versions of Artificial Neural Networks have solved a variety of challenging problems in areas such as
recognition~\cite{szegedy2013deep,graves2013speech,DCNN}, classification in high-dimensional data sets~\cite{krizhevsky2012imagenet}, image caption~\cite{XuKelvin2015,Vin2015}, machine translation~\cite{Google2016}, and generative models~\cite{Rez2014}. Due to ease of access to big data and the rapid development of adaptive artificial intelligence techniques, energy optimization and the implementation of smart cities has become a popular research trend in the energy domain. Researchers have deployed Deep Learning and Reinforcement Learning techniques in the field of energy prediction~\cite{mocanu2016deep,fan2017short} and intelligent building construction~\cite{2016intelligent}. 

Deep Learning is a sub-field of machine learning that aims to extract multilevel features by creating a functional hierarchy in which higher level features are defined based on lower level features. The difference between the Deep Learning structure and the traditional single layer fully-connected neural network is that the use of more hidden layers in the network architecture between the input layer and the output layer of the model effectively achieves more complex and nonlinear relationships. In recent years, this concept has drawn strong interests as it has become the state-of-the-art solution to solve many practical problems in several regression~\cite{xu2015regression}, classification, as well as unsupervised learning problems~\cite{lee2009convolutional,radford2015unsupervised}.

In our framework of random utility learning in a non-cooperative game setting, deep networks work as powerful models that can generalize our core model~\eqref{eq:discrete_utility} by increasing capacity and by working towards an intelligent machine learning model for predicting agent behavior. 

\subsection{Deep Neural Networks for Decision Making}

\begin{figure}[!ht]
    \subfloat[Architecture of proposed deep neural network machine learning model for decision making]{%
    \includegraphics[width=0.45\textwidth]{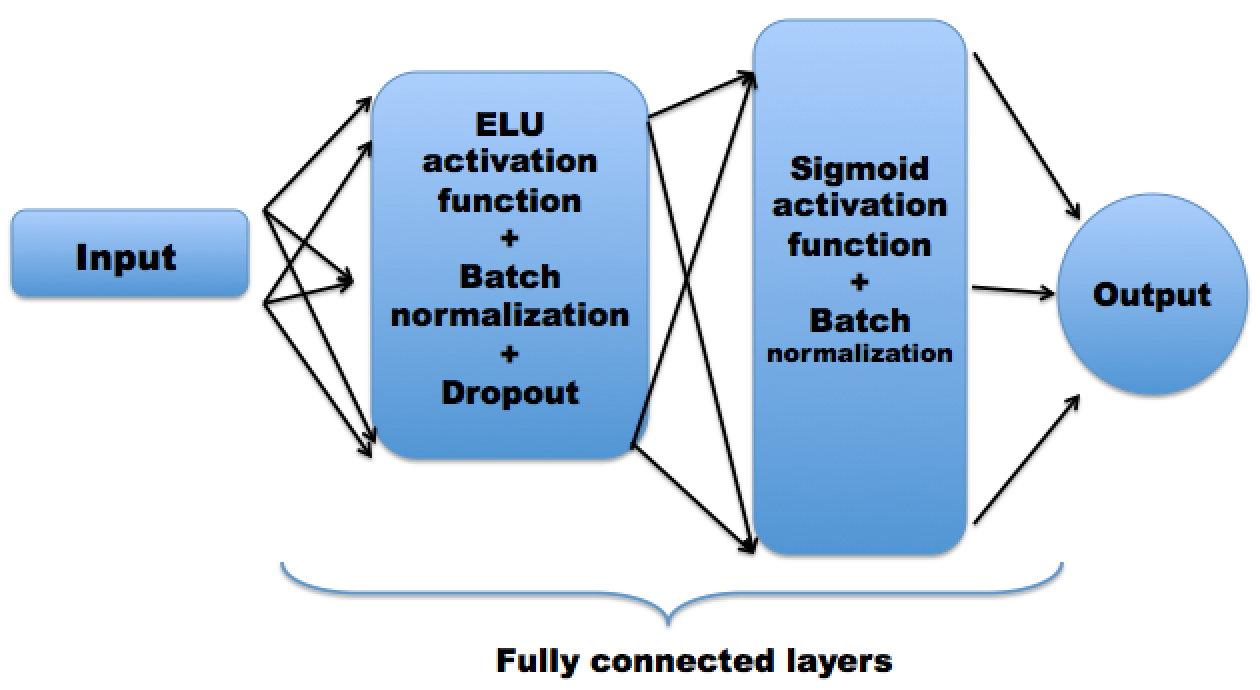}
    }
    \hfill
    \subfloat[Sequential random utility estimation leveraging deep bi-directional recurrent neural networks]{%
    \includegraphics[width=0.45\textwidth]{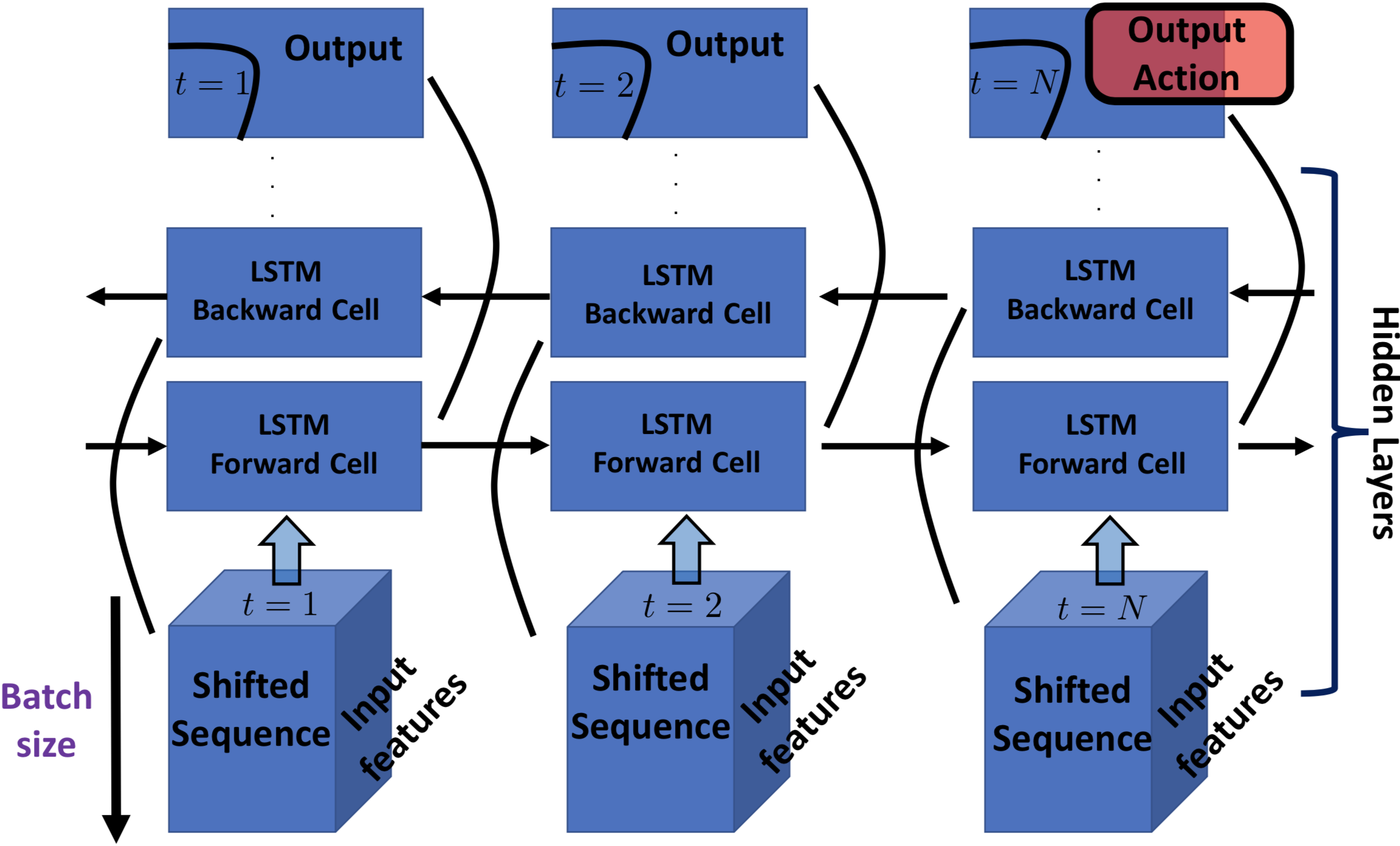}
    }
    \hfill
    \subfloat[Conventional and recurrent based deep auto-encoder]{%
    \includegraphics[width=0.45\textwidth]{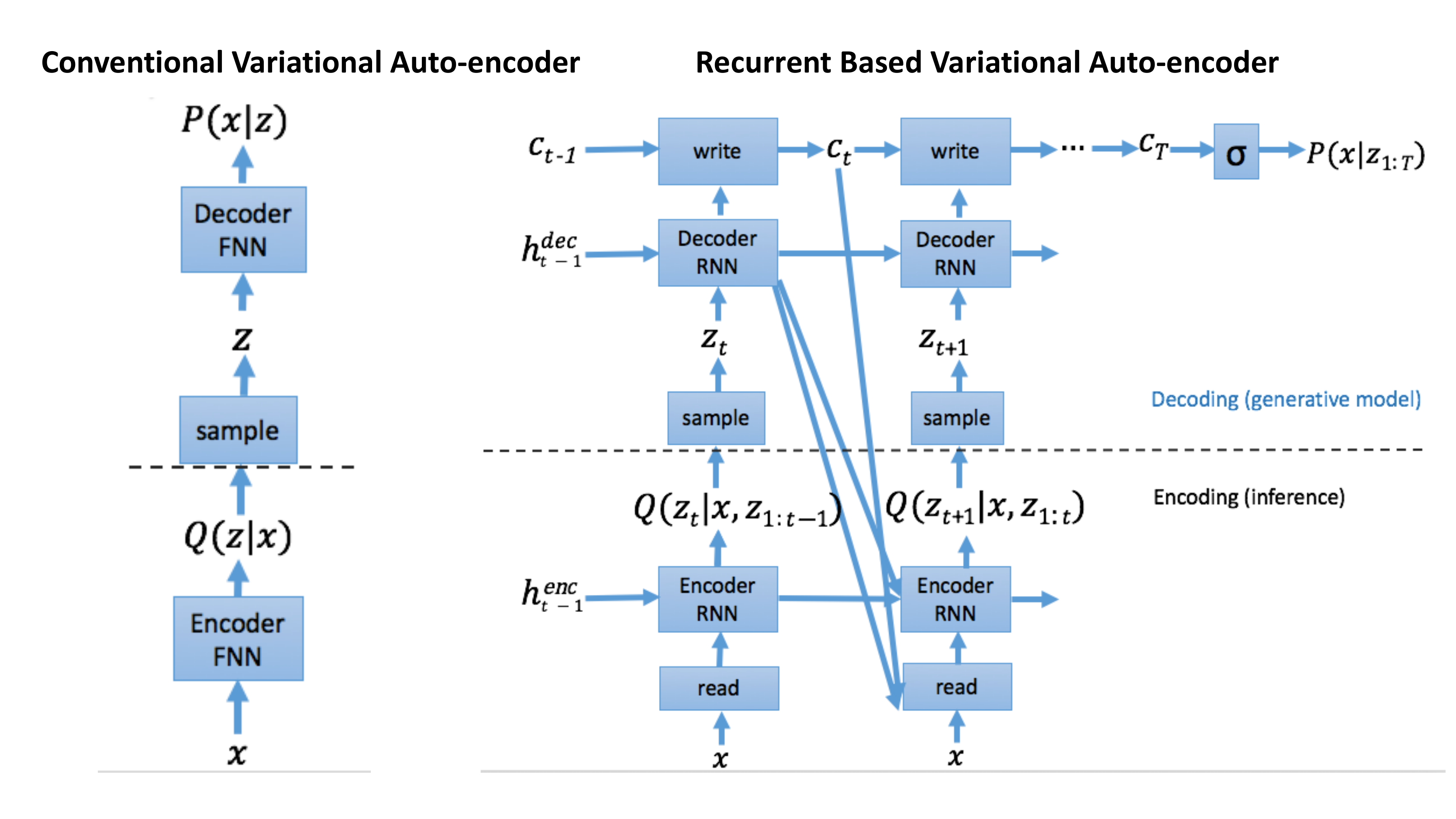}
    }
    \hfill
\caption{Proposed deep neural networks, deep bi-directional recurrent neural network and deep auto-encoders}
\label{fig:DNN_DRNN}
\end{figure}

Deep neural network techniques have drawn ever-increasing research interests ever since Deep Learning in the context of rapid learning algorithms was proposed in 2006~\cite{hinton2006fast}. Our approach has the inherent capacity to overcome deficiencies of the classical methods that are dependent on the limited series of features located in the training data set (e.g. in our setting are the features resulting from mRMR feature selection algorithm). A deep neural network can be seen as a typical feed-forward network in which the input flows from the input layer to the output layer through a number of hidden layers (in general there are more than two). Our proposed deep neural network for random utility learning is depicted in Figure~\ref{fig:DNN_DRNN}. The number of used layers can be dynamic and is a hyper-parameter that can be tuned. Usually two or three hidden layers are enough to represent a large capacity model. It is expected that a deep neural network model compared with a single hidden layer architecture achieves better performance in the classification predictive accuracy. 

Our proposed deep neural network model for random utility learning includes exponential linear units (ELUs)~\cite{goodfellow2016deep} at each hidden layer. The usage of exponential linear units (ELUs)~\cite{goodfellow2016deep} normally adds an additional hyper-parameter in the search space as a trade-off for significant increases in fitting accuracy due to enormous decrements of "dead" units --- a classical problem of rectified linear unit (ReLU) implementations~\cite{clevert2015fast}. The output layer is modeled using sigmoid units for classifying agents' discrete choices. The proposed model is optimized by minimizing the cross-entropy cost function using stochastic gradient descent combined with a Nesterov optimization scheme. The initialization of the weights utilizes He normalization~\cite{he2015delving} which gives increased performance and better training results. Unlike a random initialization, He initialization avoids local minimum points and makes convergence significantly faster. Batch Normalization~\cite{ioffe2015batch} has also been adapted in our deep neural network framework to improve the training efficiency and to address the vanishing/exploding gradient problems in the training of deep neural networks. By using Batch Normalization, we avoid drastic changes in the distribution of each layer’s inputs during training while the deep network's parameters of the previous layers keep changing. Knowing that adding more capacity in our deep neural network model will potential lead to over-fitting, we apply a dropout technique~\cite{srivastava2014dropout} as a regularization step. Using dropout, we apply a simple technique in the training phase (both in forward and backward graph learning traversal steps): each neuron, excluding the output neurons, has a probability to be totally ignored. The probability to ignore a neuron is another hyper-parameter of the algorithm and normally gets values between 50\% - 70\%.

\subsection{Deep Bi-directional Recurrent Neural Networks for Sequential Decision Making}

One of the basic drawbacks of both benchmark random utility learning models as well as the proposed deep neural networks is that they have strong assumptions for the data generation process. One important challenge for efficient learning of sequential decision making models is the actual modeling of the dependence of future actions of an agent with the present and (or) previous action(s). In particular, an agent naturally tries to co-optimize around a set of discrete choices and gains the higher utility~\eqref{eq:opt-seq}. Both benchmark models and deep neural networks adopt the assumption of \textbf{independent and identically distributed} data points. Hence, one way to model the underlying time series dependencies is through efficient feature engineering and by potentially using a novel feature selection algorithm. In Section~\ref{social_game_data_set}, we use domain knowledge along with a pooling \& picking method to create a feature set that can accurately predict agents' behavior. However, this step helps sparingly in the presence of time series dependencies and cannot generalize. 

Leveraging the latest Deep Learning models, like recurrent neural networks, we try mainly to address the issue of time dependence by looking at temporal dependencies within the data. Recurrent neural networks have the capability to allow information to persist, even over long periods, by simply inserting loops that point to them. As we see in the architecture of a deep bi-directional recurrent neural network in Figure~\ref{fig:DNN_DRNN}, information passes from one time step of the network to the next. The information of the network passes to successor nodes. In the case of a bi-directional recurrent neural network, information flows also in the opposite direction to the predecessor. In a simple implementation however, recurrent neural networks tend to either vanish or become incapable of modeling long-term dependencies. In our proposed novel sequential utility learning model, we enable an end-to-end training using Long Short Term Memory cells (LSTM). LSTM cells overcome the problem of modeling long-term dependencies as they are designed explicitly for this reason. Equations~\eqref{eq:lstm_gates},~\eqref{eq:state_up} describe exactly how LSTM cells work as an approach to modeling long term dependencies. Mainly, LSTM includes several gates that decide how long-term --- short-term relations should be modeled. The overall output of the LSTM cell is a combination of sub-gates describing the term dependencies.


\begin{eqfloat}
\begin{equation}
\begin{aligned}
i_t &= \sigma(W_{xi}x_t + W_{hi}h_{t-1} + b_i)\\
f_t &= \sigma(W_{xf}x_t + W_{hf}h_{t-1} + b_f)\\
o_t &= \sigma(W_{xo}x_t + W_{ho}h_{t-1} + b_o)
\end{aligned}
\label{eq:lstm_gates}
\end{equation}
\caption{Modeling of LSTM unit main gates.}
\end{eqfloat}

\begin{eqfloat}
\begin{equation}
\begin{aligned}
j_t &= \sigma(W_{xj}x_t + W_{hj}h_{t-1} + b_j)\\
c_t &= f_t \otimes c_{t-1} + i_t \otimes j_t\\
h_t &= \tanh(c_t) \otimes o_t
\end{aligned}
\label{eq:state_up}
\end{equation}
\caption{Input transform and state update at LSTM unit.}
\end{eqfloat}

Our data pre-processing step as well as deep bi-directional architecture is described in Figure~\ref{fig:DNN_DRNN}. Given an agent's actions, we define a time step $N$ (sliding window of actions), which is a hyper-parameter and models time series dependencies in agent actions. Each training instance in the network is a tensor with the following dimensions:

\begin{itemize}
    \item \textbf{mini-batch size:} Using mini-batch stochastic gradient descent, we select a batch size. Typically, we use some factors of time step $N$ (such as 2-3 times N).
    \item \textbf{target sequence:} Similar to a normal time series input sequence for a deep neural network architecture, it has shifted tensors with appropriate time steps. It combines the tensor with a window of past actions (i.e. sliding window).
    \item \textbf{input features:} All available features from our data set.
\end{itemize}

Our deep network is designed to leverage sequential data and build several layers and time steps of hidden memory cells (LSTMs). Moreover, we propagate the unrolled deep network both forward and backward (bi-directional recurrent neural network) for modeling the exact series time-lagged features for agents' actions. For the proposed deep bi-directional recurrent neural network, we use three hidden layers. To perform classification for the agents' actions, we add a fully connected layer containing two neurons (one per class). The fully connected layer is in turn connected to the output of the ending time-step propagating network, and this is finally followed by a soft-max layer, which actually performs the classification task. Similar to the deep neural network model, we optimize by minimizing the cross-entropy cost function using stochastic gradient descent combined with a Nesterov optimization scheme. Additionally, we employ an exponentially decaying learning rate as our learning rate schedule. Again, the initialization of the weights utilizes He normalization~\cite{he2015delving}, which gives increased performance and better training results. For dealing with the enormous capacity of the proposed deep network, we apply dropout as a regularization step.

\subsection{Deep Learning for Generative Sequential Decision Models}

On top of the existing data set resulting from our experiment, researchers can create other even larger data sets based on the existing ones. The idea of bootstrapping~\cite{Hastie09} is widely applied both in statistics and machine learning in many applications along with the creation of new data sets mimicking the original one. However, bootstrapping is not a scalable solution, and as data sets become larger and larger, the computational complexity eliminates the capabilities of the system. In our approach, we propose the use of deep variational autoencoders~\cite{kingma2013auto} as an approach to create a nonlinear manifold (encoder) that can be used as a generative model. Variational auto-encoders formalize the necessary generative model in the framework of probabilistic graphical models by maximizing a lower bound on the log-likelihood of the given high dimensional data. Furthermore, variational auto-encoders can fit large high dimensional data sets (like our social game application) and train a deep model to generate data like the original data set. In a sense, generative models automate the natural features of a data set and then provide a scalable way to reproduce known data. This capability can be employed either in the utility learning framework for boosting estimation or as a general way to create simulations mimicking occupant behavior-preferences possible in software like EnergyPlus \footnote{https://energyplus.net}

Using such a Deep Learning model, we can acquire generated samples by simply enabling the latent space of the auto-encoder and re-sampling using the decoder component. In Figure~\ref{fig:DNN_DRNN}, we provide the overall idea behind training a variational auto-encoder, which resumes as a probabilistic auto-encoder. We use two hidden layers in encoder and decoder while tying parameters between them. Also, the latent space is modeled using a Gaussian distribution. By using this architecture of deep auto-encoder however, we limit the generative model in applications in which the data process has a natural time-series dependence. Hence, we proposed the implementation of a recurrent based variational auto-encoder~\cite{gregor2015draw}. In its architecture, shown in Figure~\ref{fig:DNN_DRNN}, the proposed recurrent based variational auto-encoder allows time-series modeling for progressive refinement and spatial attention in the shifted tensor inputs. Using progressive refinement, the deep network simply breaks up the joint distribution over and over again in several steps resulting in a chain of latent variables. This gives the capability to sequentially output the time-series data rather than compute them in a single shot. Moreover, a recurrent based variational auto-encoder can potentially improve the generative process over the spatial domain. By adding time-series in the model as tensors with shifted data points, we can reduce the burden of complexity by implementing improvements over small regions of the tensor input at a time instance (spatial attention). 

Adapting those mechanisms, we achieve reduction of the complexity burden that an auto-encoder has to overcome. As a result, using a recurrent based variational auto-encoder allows for more generative capabilities that can handle larger, more complex distributions such as that in the given social game time-series. Our models were tested in several sets of data from individual occupants and was highly capable of randomly generating new data with extraordinary similarities to the training data. This fantastic result-adaptation provides an interesting tool for generating new data on top of the existing ones and provides more flexibility in the application of the data in several real scenario mechanisms like demand response. 

\section{Experimental Results}
\label{sec:res}
We now present the results of the proposed random utility learning method applied to high-dimensional IoT data collected from the network game experiment in the Fall 2017 and Spring 2018 semester.

As we previously described, our data set consists of the per-minute high-dimensional data of occupants' usage across several resources in their rooms. We evaluate the performance of random utility learning under two characteristic scenarios: a) having full-information from the installed IoT sensors for performing Step-ahead predictions. In this scenario, IoT sensors are continuously feeding information from the previous actions of the occupants. b) under this scenario, called Sensor-free, we stop taking into account the IoT sensors' readings in each room. In the second case, the rich aggregated past-historic features of the occupants are missing. For this case, we have a model in which we use only features that we can acquire from external weather conditions (e.g. from a locally installed weather station), information about occupant engagement with the web-portal, and seasonal dummy variables. All of these features are much easier to be acquired without needing to keep the highly accurate but expensive IoT devices.

We present estimation results for the complete data set in both Fall and Spring versions of the experiment for two characteristic occupants. The first occupant, in both Fall and Spring semester results, is considered a top rank player in the game with more aggressive behavior towards curtailing energy usage (e.g. in the results for this occupant in the Fall semester, there is not any usage of desk light in the testing data set---not even a minute). The second occupant, in both Fall and Spring semester results, is considered a middle rank player in the game with a behavior mixed with some energy efficient actions across several resources while also maintaining a daily usage pattern. We used the first four game periods for the training of our models:

\begin{itemize}
    \item Fall semester training/testing data set: The training data set runs from September 12th, 2017 to November 19th, 2017 (ten weeks of data). It has approximately 100,800 distinct data points per occupant with per-minute resolution. The testing period considered the next bi-weekly game, from November 20th, 2017 to December 3rd, 2017, has a total of 20,160 distinct data points per occupant.
    \item Spring semester training/testing data set: The training data set runs from February 19th, 2018 to April 22nd, 2018 (nine weeks of data). It has approximately 90,720 distinct data points per occupant with per-minute resolution. The testing period considered the next bi-weekly game, from April 23rd, 2018 to May 6th, 2018, has a total of 20,160 distinct data points per occupant.
\end{itemize}

Before we trained our benchmark classifiers, we applied the mRMR algorithm to the total data set (all occupants data) in the training period. This accounts for almost 4 million distinct data points in the Fall semester data set and 2.5 million distinct data points in the Spring semester data set. mRMR results in several top features in both the Fall and Spring semester data sets. Interestingly, mRMR included several external features in the top relevant feature candidates. In particular, the presence of external humidity as an important feature for the ceiling fan is a good demonstration of the mRMR algorithm's capability to extract salient features. Moreover, features like survey points illustrate that some occupants co-optimized their resource usage while also trying to view their point balance, usage, and ranking in the game (e.g. visiting the web-portal).



\subsection{Prediction on the Benchmark \& Deep Learning Framework Results}

For learning optimal random utility models in the benchmark setting, we use the top twenty-five resulting features from the mRMR algorithm along with a pre-processing step of SMOTE with SVM initialization. Using SMOTE, we boost the accuracy of benchmark models due to the fact that our data set was heavily imbalanced. We achieve decent accuracy using well-trained benchmark models. Especially in the Step-ahead predictions, all of the classifiers achieve decent AUC scores in both the Fall and Spring semester results, as shown in Tables~\ref{tab:auc_fall_step_ahead}, \ref{tab:auc_spring_step_ahead}. In the Sensor-free results, Tables~\ref{tab:auc_fall_sensor_free}, \ref{tab:auc_spring_sensor_free}, we have a significant drop in the achieved accuracy, but this is expected given that the IoT feed is decoupled. However, even in Sensor-free examples we are able to predict occupants' behavior using less representative features and having excluded the "costly" IoT sensors.

For the Deep Learning models, we used the exact same data set. For the deep neural networks, we used training data resulting from the applied SMOTE step as in the benchmark analogy. We used two hidden layers of the feed-forward neural network, with 50\% dropout and stochastic gradient descent method leveraging Nesterov's Momentum to accelerate convergence.  

To further exploit the continuity of the sequential decision making model, we experiment on the bi-directional deep recurrent neural network. We used a time sliding window---time step of two hours (120 distinct data-points). We processed the data without being pre-processed from the SMOTE algorithm as we wanted to keep control of the underlying sequence of actions of the occupants (temporal dependences of the data). We used three hidden layers with 60\% dropout rate and we applied exponentially decaying learning rate as our learning rate schedule. In the training of bi-directional recurrent neural networks, we applied the principle of early stopping using a validation data set (small portion of the data --- one week period) over the AUC metric. For our deep bi-directional networks, thirty-five epochs were optimal to be trained. 

To evaluate the effectiveness of our proposed deep learning framework, we present the AUC scores of a representative example for comparison. From the results, it is clear that deep RNN outperforms the majority of alternative algorithms with few outliers (bold highlighted scores). One important remark is that deep RNN performs best even when compared to Random Forest, which is considered a top robust performing classification model. Interestingly, deep NN could achieve better performance in some examples over the Random Forest classifier, but this is not a general case. 


\begin{table}[h!]
\centering
\begin{tabular}[!h]{ |c|c|c|c| }
 \hline
 Occupant $1$ / $2$ & Ceiling Fan & Ceiling Light & Desk Light\\ 
 \hline
Logistic regression & 0.74 / 0.83 & 0.75 / 0.78 & N/A / 0.78\\ 
 \hline
Penalized $l1$ Logistic regression & 0.75 / 0.80 & 0.75 / 0.77 & N/A / 0.78\\ 
\hline
Bagged Logistic regression & 0.76 / 0.84 & 0.77 / 0.80 & N/A / 0.79 \\
 \hline
LDA &  0.75 / 0.81 & 0.75 / 0.78 & N/A / 0.74\\
 \hline
K-NN &  0.70 / 0.76 & 0.72 / 0.77 & N/A / 0.74 \\
 \hline
Support Vector Machine &  0.74 / 0.82 & 0.76 / 0.78 & N/A / 0.76\\
 \hline
 Random Forest & 0.92 / 0.91 & 0.79 / 0.78 & N/A / 0.98\\
 \hline
 Deep Neural Network & 0.82 / 0.80 & 0.80 / 0.76 & N/A / 0.78 \\
 \hline
 Deep Bi-directional RNN & \textbf{0.93} / \textbf{0.97} & \textbf{0.89} / \textbf{0.85} & N/A / \textbf{0.99} \\
 \hline
\end{tabular}
\caption{AUC scores using Fall semester data of two representative occupants towards Step-ahead predictions.}
\label{tab:auc_fall_step_ahead}
\end{table}

\begin{table}[h!]
\centering
\begin{tabular}[!h]{ |c|c|c|c| }
 \hline
 Occupant $1$ / $2$ &Ceiling Fan & Ceiling Light & Desk Light\\ 
 \hline
Logistic regression &  0.62 / 0.65 & 0.61 / 0.61 & N/A / 0.68\\
 \hline
Penalized $l1$ Logistic regression &  0.59 / 0.65 & 0.55 / 0.56 & N/A / 0.64\\
\hline
Bagged Logistic regression &  0.64 / 0.66 & 0.61 / 0.59 & N/A / 0.68\\
 \hline
LDA &  0.63 / 0.65 & 0.60 / 0.58 & N/A / 0.68\\
 \hline
K-NN &  0.56 / 0.53 & 0.50 / 0.56 & N/A / 0.55 \\
 \hline
Support Vector Machine &  0.64 / 0.65 & 0.60 / 0.60 & N/A / 0.68\\
 \hline
 Random Forest & 0.58 / 0.60 & 0.56 / 0.59 & N/A / 0.63\\
 \hline
 Deep Neural Network &  0.59 / 0.55 & 0.53 / 0.60 & N/A / 0.64\\
 \hline
 Deep Bi-directional RNN &  \textbf{0.69} /\textbf{ 0.71} & \textbf{0.65} / \textbf{0.66} & N/A / \textbf{0.76}\\
 \hline
\end{tabular}
\caption{AUC scores using Fall semester data of two representative occupants towards Sensor-free predictions.}
\label{tab:auc_fall_sensor_free}
\end{table}


\begin{table}[h!]
\centering
\begin{tabular}[!h]{ |p{1.75cm}|p{1.5cm}|p{1.25cm}|p{1.75cm}|p{1.5cm}| }
 \hline
 Occupant $1$ / $2$ &Ceiling Fan & Air-con & Ceiling Light & Desk Light\\ 
 \hline
Logistic regression & 0.71 / 0.84  & 0.76 / 0.82 & 0.75 / 0.83 & 0.76 / 0.73\\ 
 \hline
Penalized $l1$ Logistic regression & 0.71 / 0.84 & 0.76 / 0.82 & 0.75 / 0.83 & 0.76 / 0.71\\ 
\hline
Bagged Logistic regression & 0.73 / 0.85 & 0.73 / 0.83 & 0.76 / 0.84 & 0.79 / 0.74 \\
 \hline
LDA & 0.70 / 0.87 & 0.73 / 0.83 & 0.75 / 0.83 & 0.70 / 0.92 \\
 \hline
K-NN &  0.70 / 0.84 & 0.76 / 0.86 & 0.68 / 0.81 & 0.73 / 0.76 \\
 \hline
Support Vector Machine &  0.70 / 0.86 & 0.75 / 0.83 & 0.75 / 0.83 & 0.70 / 0.49 \\
 \hline
 Random Forest & 0.83 / \textbf{0.99} & 0.83 / 0.81 & \textbf{0.99} / \textbf{0.98} & 0.96 / 0.87 \\
 \hline
 Deep Neural Network & 0.74 / 0.86 & 0.78 / 0.87 & 0.77 / 0.84 & 0.84 / 0.90 \\
 \hline
 Deep Bi-directional RNN & \textbf{0.91} / 0.98 & \textbf{0.89} / \textbf{0.94} & \textbf{0.99} / 0.97 & \textbf{0.99} / \textbf{0.95} \\
 \hline
\end{tabular}
\caption{AUC scores using Spring semester data of two representative occupants towards Step-ahead predictions.}
\label{tab:auc_spring_step_ahead}
\end{table}

\begin{table}[h!]
\centering
\begin{tabular}[!h]{ |p{1.25cm}|p{1.25cm}|p{1.25cm}|p{1.25cm}|p{1.25cm}| }
 \hline
 Occupant $1$ / $2$ &Ceiling Fan & Air-con & Ceiling Light & Desk Light\\ 
 \hline
Logistic regression & 0.55 / 0.71  & 0.73 / 0.61 & 0.55 / 0.69 & 0.50 / 0.70\\ 
 \hline
Penalized $l1$ Logistic regression & 0.55 / 0.72 & 0.70 / 0.61 & 0.55 / 0.70 & 0.50 / 0.73\\ 
\hline
Bagged Logistic regression & 0.54 / 0.73 & 0.73 / 0.73 & 0.54 / 0.71 & 0.51 / 0.65 \\
\hline
LDA & 0.55 / 0.71 & 0.73 / 0.68 & 0.55 / 0.70 & 0.51 / 0.59 \\
 \hline
K-NN &  0.50 / 0.63 & 0.57 / 0.69 & 0.54 / 0.69 & 0.57 / 0.50 \\
 \hline
Support Vector Machine &  0.55 / 0.70 & 0.73 / 0.67 & 0.55 / 0.70 & 0.50 / 0.71 \\
 \hline
 Random Forest & 0.58 / 0.66 & 0.65 / 0.54 & 0.54 / 0.68 & 0.50 / 0.50 \\
 \hline
 Deep Neural Network & 0.56 / 0.67 & 0.68 / 0.53 &  0.54 / 0.67  & 0.50 / 0.50 \\
 \hline
 Deep Bi-directional RNN & \textbf{0.66} / \textbf{0.79} & \textbf{0.80} / \textbf{0.78} & \textbf{0.64} / \textbf{0.77} & \textbf{0.62} / \textbf{0.83} \\
 \hline
\end{tabular}
\caption{AUC scores using Spring semester data of two representative occupants towards Sensor-free predictions.}
\label{tab:auc_spring_sensor_free}
\end{table}

In Table~\ref{tab:autoencoder_res}, we present the results of two trained generative models using the full data set of a randomly selected occupant in the Fall semester. We trained both a conventional auto-encoder and a recurrent based auto-encoder. In Table~\ref{tab:autoencoder_res}, we present several selected features, either from the interior of a dorm room or external weather data. For the evaluation of the artificially generated time-series using the proposed auto-encoders, we utilize dynamic time warping (DTW) for measuring the similarity between the two temporal sequences --- the ground truth and the artificial data from the generative model. Dynamic time warping (DTW) is an extension of Levenshtein distance and can be computed in pseudo-polynomial time~\cite{salvador2007toward}. In bold, we see that the recurrent based auto-encoder achieves a smaller DTW score in most of the features, leading to a generative model that isn't mimicking exactly or is way too different from the original data set. Wanting to evaluate the statistical significance of the calculated DTW scores from the recurrent based auto-encoder, we used a permutation hypothesis test. In this approach, we permute original and generated time-series and we computed their DTW score looking for events that are more "extreme" than the one that is presented in Table~\ref{tab:autoencoder_res}. Interestingly, we have inside and outside weather based features (temperature and humidity) that have zero p-values showing that the DTW score using a recurrent based auto-encoder are significant. For indoor device status features however, p-values are large, showing that the DTW score has high variability under the permutation test.

\begin{table}[h!]
\centering
\begin{tabular}[!h]{ |p{3.5cm}|p{1.75cm}|p{1.5cm}|p{0.5cm}|}
\hline
Time Series Feature  & Conventional Auto-encoder & RNN-based Auto-encoder & p-values \\
\hline
Ceiling Fan Status (On / Off)          &  1.5e+04 &  \textbf{1.2e+04} &  0.11 \\
\hline
Ceiling Light Status (On / Off)        &  1.6e+04 &  \textbf{2.2e+03} &  1.0 \\
\hline
Desk Light Status (On / Off)           &  6.7e+03 &  \textbf{0.0e+00} &  1.0 \\
\hline
Dorm Room Temperature                &  1.3e+05 &  \textbf{1.2e+05} &  0.0 \\
\hline
Dorm Room Humidity            &  4.8e+05 & \textbf{3.7e+05}
 &  0.0 \\
\hline
External Temperature               &  1.0e+05 &  1.8e+05 &  0.0 \\
\hline
External Humidity          &  2.9e+05 &  4.3e+05 &  0.0 \\
\hline
\end{tabular}
\caption{DTW score --- feature comparison between proposed generative models (autoencoders).}
\label{tab:autoencoder_res}
\end{table}

\subsection{Energy Savings}

Here we present achieved savings in both Fall and Spring semester version of the social game. For quantifying the results, we employ hypothesis testing (A/B testing) using dorm occupants' usage data before and after the beginning of the experiment. Hypothesis testing is a standard technique of great importance used across all fields of research. In the energy domain, we see many examples where testing is crucial in determining the feasibility of the hypothesis. As an example, testing energy-GDP causality has resulted in a lot of disparate results, largely because of omitted variable bias and the lack of a priori hypothesis testing \cite{liddle2015revisiting}. 


In Ang’s paper, we see that hypothesis testing is central in determining the correlation between CO2 emissions and energy consumption in France \cite{ang2007co2}. Multiple tests are performed to find causal links of output energy and pollution. This is also complemented with Granger non-causality tests and exogeneity tests, for comparison between the short run and the long run. This is an idea that we could incorporate, since one of our goals is to provide a suitable long run forecast. Then, we would have a more informed idea of what we can expect in different time periods. This suggests a natural extension to our work in this paper, where we can further analyze the sequential discrete game to account for time-series dependencies. Additionally, hypothesis testing is frequently supplemented with other statistical procedures such as cross-validation and information criteria. This is relevant, since in this paper, we have attempted to predict and improve forecasting performance using various Deep Learning techniques. 

In Tables~\ref{tab:fall_hyp} and~\ref{tab:spring_hyp}, we see the hypothesis testing values for the different devices in both iterations of the experiment (Fall and Spring). 
In the tables, the Before column denotes the data points gathered from before the game was officially started. The After column is the data during the game period. Data points in the tables are bucketed in both weekday and weekend data and represent the average usage of all the occupants. Usage is defined in minutes per day. In all cases of the devices, we have a significant drop in usage between the two periods. Drop in usage is given in the column named $\Delta$ \%, and indicates reduction in the average usage of all the participating occupants. The p-values resulting from the 2-sample t-tests show that the change in usage patterns is highly significant. Moreover, we can see that a much larger drop in usage is achieved over the weekends. Lastly, in Table~\ref{tab:tod_hyp} we compare the occupants usage patterns after the beginning of the game. In particular, we perform 2-sample t-tests, which in most of the cases show that there is a significant difference in occupants' usage patterns between weekdays and weekends. These are significant results showing how we can optimally incentivize occupants in residential buildings to reduce energy usage, especially over weekend periods. 

\begin{table}[h!]
\centering
\begin{tabular}[!h]{ |p{0.7cm}|p{0.7cm}|p{0.7cm}|p{0.7cm}|p{0.7cm}|p{0.7cm}|p{0.7cm}|p{0.7cm}|p{0.7cm}| }
\hline 
   & \multicolumn{4}{|c|}{Weekday} & \multicolumn{4}{|c|}{Weekend}\\
 \hline
 Device & Before (Mean) & After (Mean) & $p$-value & $\Delta$ \% & Before (Mean) & After (Mean)  & $p$-value & $\Delta$ \%\\
 \hline
Ceiling Light & 417.5 & 393.9 & 0.02 & 5.6 & 412.3 & 257.5 & 0 & 37.6\\
 \hline
 Desk Light & 402.2 & 157.5 & 0 & 60.8  & 517.6 & 123.3 & 0 & 76.2\\
 \hline
Ceiling Fan & 663.5 & 537.6 & 0 & 19.0  & 847.1 & 407.0 & 0 & 51.9\\ 
\hline
\end{tabular}
\caption{Fall Game (Before vs After) usage hypothesis testing. }
\label{tab:fall_hyp}
\end{table}

\begin{table}[h!]
\centering
\begin{tabular}[!h]{ |p{0.65cm}|p{0.65cm}|p{0.65cm}|p{0.4cm}|p{0.3cm}|p{0.65cm}|p{0.65cm}|p{0.4cm}|p{0.3cm}| }
 \hline 
   & \multicolumn{4}{|c|}{Weekday} & \multicolumn{4}{|c|}{Weekend}\\
 \hline 
 Device & Before (Mean) & After (Mean) & $p$-value & $\Delta$ \% & Before (Mean) & After (Mean)  & $p$-value & $\Delta$ \% \\
 \hline
Ceiling Light & 452.0 & 314.2 & 0 & 30.5 & 426.0 & 195.6 & 0 & 54.1\\ 
 \hline
 Desk Light & 430.1 & 104.6 & 0 & 75.7 & 509.4 & 81.5 & 0 & 84\\
 \hline
Ceiling Fan & 777.4 &  541.6  & 0 & 30.3 & 847.1 & 331.8 & 0 & 60.8\\ 
\hline
 Air Con & 469.8 & 225.8 & 0 & 51.9 & 412.3 & 81.8 & 0 & 80.2\\
 \hline
\end{tabular}
\caption{Spring Game (Before vs After) usage hypothesis testing. }
\label{tab:spring_hyp}
\end{table}
 
\begin{table}[h!]
\centering
\begin{tabular}[!h]{ |p{0.9cm}|p{0.75cm}|p{0.75cm}|p{0.75cm}|p{0.75cm}|p{0.75cm}|p{0.75cm}| }
\hline 
   & \multicolumn{3}{|c|}{Spring} & \multicolumn{3}{|c|}{Fall}\\
 \hline
 Device & Wday (Mean) & Wkend (Mean) & $p$-value & Wday (Mean) & Wkend (Mean) & $p$-value \\
 \hline
Ceiling Light & 314.2 & 195.6 & 0 & 393.9 & 257.5 & 0\\ 
 \hline
 Desk Light & 104.6 & 81.5 & 0.1 & 157.5 & 123.3 & 0.01 \\
 \hline
Ceiling Fan & 541.6 &  331.8  & 0 & 537.6 &  407.0  & 0\\ 
\hline
 Air Con & 225.8 & 81.9 & 1.0 & N/A & N/A & N/A\\
 \hline
\end{tabular}
\caption{Weekday vs. Weekend usage hypothesis testing }
\label{tab:tod_hyp}
\end{table}

\subsection{Survey Results}

In this last section, we present the results of the given surveys in the Spring semester. Surveys were administered to occupants every two days and included several questions using a 5-point Likert-scale survey format. For each question's target set, we phrased it in opposite terms and we randomly sorted the questions order at each instance of survey administration. Wanting to test occupants' internal consistency regarding survey results, we deployed a Cronbach's $\alpha$ statistic test. As we see in Table~\ref{tab:cronbach}, people are internally consistent regarding their satisfaction for the lighting \& HVAC condition and provided incentives. We also note that the Cronbach's $\alpha$ is negative for the 3rd and 4th entries, regarding the awareness of energy-saving techniques and the web portal interface. This is because there are weak correlations between the variables, which is counter intuitive seeing that the questions were designed such that consistent answers were encouraged. This result suggests that dorm occupants felt that the two questions in each bucket for the third and fourth categories were not actually asking the same question in essence.

\begin{table}[]
\begin{tabular}{|p{0.1cm}|p{4.8cm}|p{0.25cm}|p{0.05cm}|}
\hline
& \multicolumn{1}{c|}{\textbf{Questions}}                                                         & \multicolumn{1}{c|}{\textbf{Cronbach's $\alpha$}} & \multicolumn{1}{c|}{\textbf{$p$-value}} \\ \hline
\textbf{\rotatebox[origin=c]{90}{Light}}      & \begin{tabular}[c]{@{}l@{}}I am satisfied with today's lighting \\ conditions.\\ Today's lighting conditions were \\ uncomfortable.\end{tabular}                                                     & 0.75                                              & 0.003                                   \\ \hline
\textbf{\rotatebox[origin=c]{90}{ Incentives }} & \begin{tabular}[c]{@{}l@{}}I am happy with the current \\ incentives provided.\\ The current incentives are \\ not satisfactory.\end{tabular}                                                        & 0.82                                              & 0.002                                   \\ \hline
\textbf{\rotatebox[origin=c]{90}{ Energy }}     & \begin{tabular}[c]{@{}l@{}}I am aware of energy-saving \\ techniques that I can use.\\ I feel unable to save energy \\ through my actions\end{tabular}                                               & \ -0.40                                             & 0                                       \\ \hline
\textbf{\rotatebox[origin = c]{90}{ Web Portal }} & \begin{tabular}[c]{@{}l@{}}I am satisfied with the current \\ web interface.\\ The web portal leaves much \\ to be desired.\end{tabular}                                                             & \ -0.24                                             & 0                                       \\ \hline
\textbf{\rotatebox[origin = c]{90}{ HVAC }}       & \begin{tabular}[c]{@{}l@{}}I am satisfied with today's HVAC \\ (thermo comfort aircon/ fan) conditions.\\ Today's HVAC (thermo comfort -- \\ aircon/fan) conditions were uncomfortable.\end{tabular} & 0.78                                              & 0.003                                   \\ \hline
\end{tabular}
\caption{Cronbach's $\alpha$ Testing for 5-point Likert-scale Survey Responses}
\label{tab:cronbach}
\end{table}

\section{Discussion, Conclusion $\&$ Future Remarks}
\label{sec:disc}
We presented a general framework for random utility learning in sequential decision-making models. We leveraged several Deep Learning architectures and proposed a novel sequential Deep Learning classifier model, which utilizes bi-directional recurrent networks along with LSTM cells. We also introduced a framework that serves as a base for creating generative models ideal for modeling human-centric architectures. On top of that, we developed a detailed random utility pipeline for several classical benchmark models. The latter is important for having a variety of models to characterize occupants' behavior, but also to compare our proposed Deep Learning models. 

To demonstrate the random utility learning methods, we applied them to data collected from a smart building social game we conducted where occupants optimized concurrently their room's resources and participated in a lottery. We were able to estimate several agent profiles and significantly reduce the forecasting error compared to all benchmark models. The deep sequential random utility learning framework outperformed all the models, and, in specific examples, it improved prediction accuracy to an extraordinary degree. This last result shows that a Deep Learning architecture that handles a sequential data process boosts the overall accuracy. Although we apply the method specifically to smart building social game data, it can be applied more generally to scenarios with the task of inverse modeling of competitive agents, and it provides a useful tool for many smart infrastructure applications where learning decision making behavior is crucial. 

By running this gaming experiment on the Nanyang Technological University campus in both the Fall 2017 (September 12th - December 3rd) and Spring 2018 (February 19th - May 6th) semesters, we enabled the gathering of a large data set including not only occupants' gaming actions but also their resource usage, occupancy preferences, and interactions with web portal. After analyzing and formulating this data set, we designed a demo web portal for the demonstration of our infrastructure and for downloading de-identified high dimensional data sets\footnote{\textit{smartNTU} demo web-portal: https://smartntu.eecs.berkeley.edu}. Our provided high-dimensional data set can serve either as a benchmark for discrete choice model learning schemes or as a great source for analyzing occupant resource usage in residential buildings. Other researchers now have the ability to easily demonstrate gaming data in a discrete choice setting, run simulations including occupants dynamic preferences (data set has one minute resolution), test correlations of actions vs external parameters like weather (e.g. we provide various weather metrics), and leverage temporal data sets in several demand response program approaches~\cite{jin2018microgrid,jin2017mod}.

\section{Acknowledge}
\label{sec:ackow}
The authors would like to thank Chris Hsu, the applications programmer at CREST laboratory, who developed and deployed the web portal application as well as the social game data pipeline architecture. Also, we want to thank Energy Research Institute (ERI@N) at Nanyang Technological University. Geraldine Thoung, Patricia Alvina and Nilesh Y. Jadhav at ERI@N kindly supported and helped during the social game experiment. This project could not succeed without the important support of our interns at Nanyang Technological University: Ying Xuan Chua, Shuen Hwee Yee, Keryn Kan, Cyndi Shin Yi Teh, Shu Yu Tan, and Yu Jie Lee. Our intern team helped in several challenges before and during the deployment of our social game. Undoubtedly, their support was highly important for the experiment and its success. This work was supported by the Republic of Singapore’s National Research Foundation through a grant to the Berkeley Education Alliance for Research in Singapore for the Singapore–Berkeley Building Efficiency and Sustainability in the Tropics (SinBerBEST) Program. The work of I. C. Konstantakopoulols was supported by a scholarship of the Alexander S. Onassis Public Benefit Foundation.

\bibliographystyle{IEEEtran}
\bibliography{journal_utility}

\end{document}